# Single-condition neural solvers encode transferable response spaces for parametric differential equations

Wenbo Cao[a,b], Weiwei Zhang[c,d,e,*]

[a] *Institute of AI for Industries, Chinese Academy of Sciences, Nanjing 211135, China*

[b] *Institute of Computing Technology, Chinese Academy of Sciences, Beijing 100190, China*

[c] *School of Aeronautics, Northwestern Polytechnical University, Xi'an 710072, China*

[d] *International Joint Institute of Artificial Intelligence on Fluid Mechanics, Northwestern Polytechnical University, Xi'an 710072, China*

[e] *National Key Laboratory of Aircraft Configuration Design, Xi'an 710072, China*

* *Corresponding author. E-mail address: aeroelastic@nwpu.edu.cn*

**Abstract.** Operator learning for parametric partial differential equations (PDEs) typically builds global models over prescribed domains, requiring cross-condition data or costly physics-constrained training. Here we show that the output Jacobian of a neural solution model trained at one condition defines a reusable response space for cross-condition solution variations. We introduce Linearized Subspace Transfer (LST) to exploit this space and recover target solutions by minimizing the target PDE-system residual over response-space coordinates. Because any single response space has finite coverage, Active Transfer Modeling (ATM) uses post-transfer residuals as coverage indicators to selectively acquire response spaces from additional single-condition models. Across six systems, single-condition response spaces supported cross-condition transfer, with enrichment improving accuracy when added spaces expanded representation capacity. Relative to evaluated physics-informed operator baselines, ATM reduced error and offline construction cost, with orders-of-magnitude accuracy gains in representative cases and millisecond-to-second target adaptation. These results establish neural solvers as reusable local parametric models.

## Introduction

Efficient solution of parametric differential equations is central to scientific machine learning and computational engineering, where repeated evaluations across varying initial and boundary conditions, forcing functions, material properties, geometries and physical parameters can render case-by-case simulation prohibitively expensive[1-3]. Operator learning addresses this challenge by approximating mappings from problem specifications directly to solution fields[4]. DeepONet[5], Fourier neural operators[6], and their physics-informed variants[7,8] have established a powerful global modeling paradigm in which a single model is constructed over a prescribed condition domain and subsequently queried throughout that domain. Data-driven approaches build such mappings from high-fidelity solutions sampled across conditions[3-6,9-11], whereas physics-informed approaches reduce reliance on labeled solutions by enforcing the governing equations during training, but shift the construction cost to residual evaluation and optimization across condition samples and spatiotemporal collocation points[7,8,12-14]. Both approaches therefore construct parametric coverage upfront through training across the condition domain, trading this offline construction for rapid amortized inference after training.

An alternative to constructing parametric coverage upfront would be to reuse neural solution models trained at individual conditions across conditions. Such models, however, are typically treated as condition-specific approximations rather than as parametric models. Yet a trained network defines more than its condition-specific prediction: its output Jacobian maps network-parameter perturbations to structured variations of the predicted field. Whether the resulting response space can represent solution changes induced by other conditions, despite the model having never been trained across the condition family, remains largely unexplored. This raises a basic question: can a neural solution model trained at a single condition encode a reusable local representation of the surrounding parametric solution family?

Several existing paradigms address related aspects of adaptation or local approximation. Transfer learning and parameter-efficient adaptation modify pretrained neural models at the target condition, typically through partial or low-rank parameter updates[15-18]. Meta-learning uses experience across related tasks to learn representations or initializations that can be rapidly adapted to new problems[19-23], whereas active learning selectively acquires informative conditions or training samples to improve coverage of the parameter domain[24-26]. Projection-based reduced-order methods, including localized and adaptively enriched variants, construct parameter-dependent trial spaces from solution snapshots[2,27-29], while randomized-feature PDE solvers operate in prescribed, untrained feature spaces[30,31]. Neural tangent kernel and linearized-network analyses establish the expressive role of Jacobian-induced tangent features and their connection to wide-network learning dynamics and function spaces[32-34]. Evolutional deep neural networks and Neural Galerkin schemes have exploited network-induced tangent spaces for sequential PDE evolution[35,36]. Closest to the present work, Linearized Subspace Refinement (LSR) showed that the output Jacobian of a trained network provides directions for correcting approximation error for the same differential-equation instance[37]. What remains unexplored is whether a neural solver constructed for one condition can instead be reused through its response space as a local parametric model for other conditions.

Here we show that the output Jacobian of a neural solution model trained at a single condition defines a reusable local response space that captures solution variations across conditions. We introduce Linearized Subspace Transfer (LST) to exploit this structure, recovering unseen-condition solutions by minimizing the full target PDE-system residual within a fixed affine response space extracted from the trained model. This representation exposes a structured least-squares problem over response coordinates and allows the response modes and their PDE-relevant derivatives to be precomputed and reused across targets. Because any single response space has finite coverage, we further introduce Active Transfer Modeling (ATM), which incrementally enriches a library of single-condition response spaces. Post-transfer residuals serve as practical coverage indicators for acquiring new models in poorly represented regions, allowing parametric coverage to be extended on demand rather than constructed globally in advance.

We evaluate this framework across six differential-equation systems spanning linear and nonlinear problems, steady and unsteady regimes, function-valued conditions and up to two spatial dimensions. The experiments establish that response spaces extracted from independently trained single-condition models support reproducible cross-condition transfer, while additional controls examine the robustness of this response structure and the rapid target adaptation enabled by LST. We further show that transferability alone does not determine the benefit of enrichment: a single source may already provide broad coverage, whereas active enrichment yields substantial gains when newly acquired sources contribute complementary representation capacity. Comparisons with evaluated physics-informed operator baselines characterize the accuracy–cost trade-off of this local-response paradigm, combining lower offline construction costs and higher accuracy with efficient but target-specific online adaptation. Together, these results establish neural solution models as reusable local parametric models rather than merely condition-specific approximations.

# Results

## Method overview

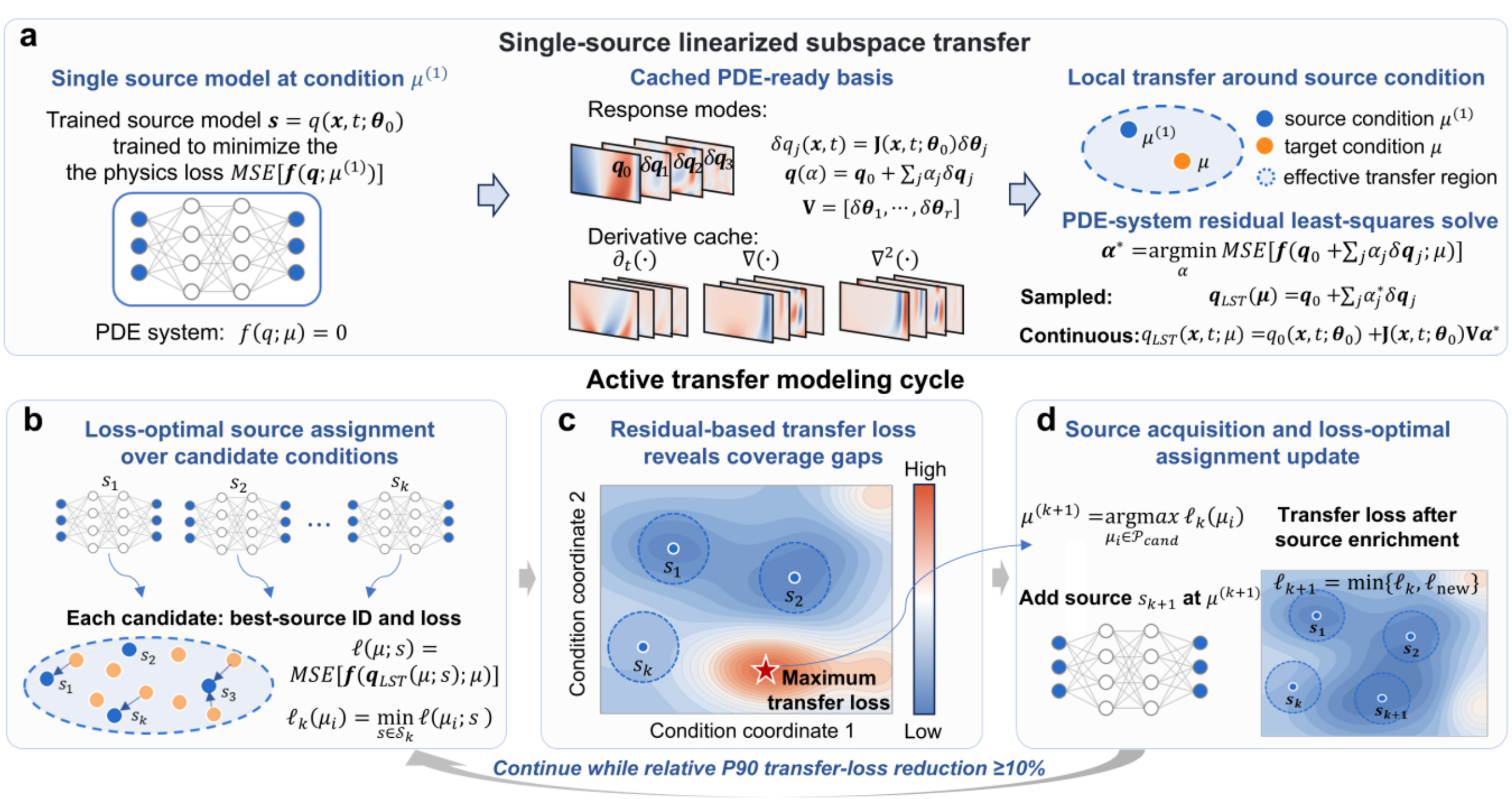

**Fig. 1 | From single-condition response spaces to active transfer modeling. a**, Linearized Subspace Transfer (LST) converts a neural solution model trained at a source condition into cached Jacobian-induced response modes and their PDE-relevant derivatives. The target solution is recovered by minimizing the target PDE-system residual over the

modal coefficients. **b**, Each candidate condition is assigned to the source response space yielding the lowest post-transfer loss. **c**, The resulting minimum residual defines the transfer loss and serves as a practical indicator of transfer coverage in the current source library. **d**, ATM trains a new source model at the maximum-loss condition and updates each candidate by comparing the new response space with its incumbent lowest-loss source. Enrichment continues while the relative reduction in the 90th percentile of transfer loss remains at least 10%.

A neural solution model $q(\boldsymbol{x},t;\boldsymbol{\theta}_0)$ trained at a single source condition $\mu^{(1)}$ provides the starting point for cross-condition transfer (Fig. 1a). We extract dominant directions of the source-model output Jacobian $\mathbf{J}$ using a matrix-free randomized SVD and form the corresponding Jacobian-induced field responses, $\mathbf{\Phi} = \mathbf{JV}$, as reusable response modes. Together with the source prediction $q_0$, these modes define a rank-truncated affine response space. Because a source may be reused across multiple targets, the response modes and their PDE-relevant spatial and temporal derivatives are precomputed and cached.

Linearized Subspace Transfer (LST) seeks the target solution in this affine space $q_{\mathrm{LST}}(\boldsymbol{x},t) = q_0(\boldsymbol{x},t) + \mathbf{\Phi}(\boldsymbol{x},t)\boldsymbol{\alpha}$. The target-specific response coordinates $\boldsymbol{\alpha}$ are determined by minimizing the target PDE-system residual. Although this residual may be nonlinear in $\boldsymbol{\alpha}$, the field and its required derivatives remain affine in the response coordinates, allowing structure-exploiting least-squares updates to be assembled directly from the cached representation without re-evaluating or differentiating the source network.

Because any single source response space has finite coverage, Active Transfer Modeling (ATM) extends LST by incrementally enriching a library of source-specific response spaces (Fig. 1b–d). For each candidate condition, ATM stores the lowest post-transfer loss and its associated source, using this residual-based loss as a practical coverage indicator. The largest-loss candidate is selected for source acquisition, after which the cached losses and assignments are updated wherever the new source performs better. For an unseen target, ATM routes to the source assigned to its nearest candidate and performs one LST solve. Enrichment stops when the relative reduction in the 90th-percentile candidate loss falls below 10%.

**Trained Jacobian response spaces enable single-source transfer**

We first isolate single-source transfer on the one-dimensional viscous Burgers benchmark of Wang et al.[7], using the same governing equation and initial-condition family. A network trained for one source initial condition provides cached response modes that are reused to approximate solutions for unseen target initial conditions with markedly different spatial profiles (Fig. 2c). For each target, LST determines the response coordinates solely by minimizing the target PDE-system residual, testing whether the output tangent space of a single-condition model can represent condition-induced solution variations without reference-solution data.

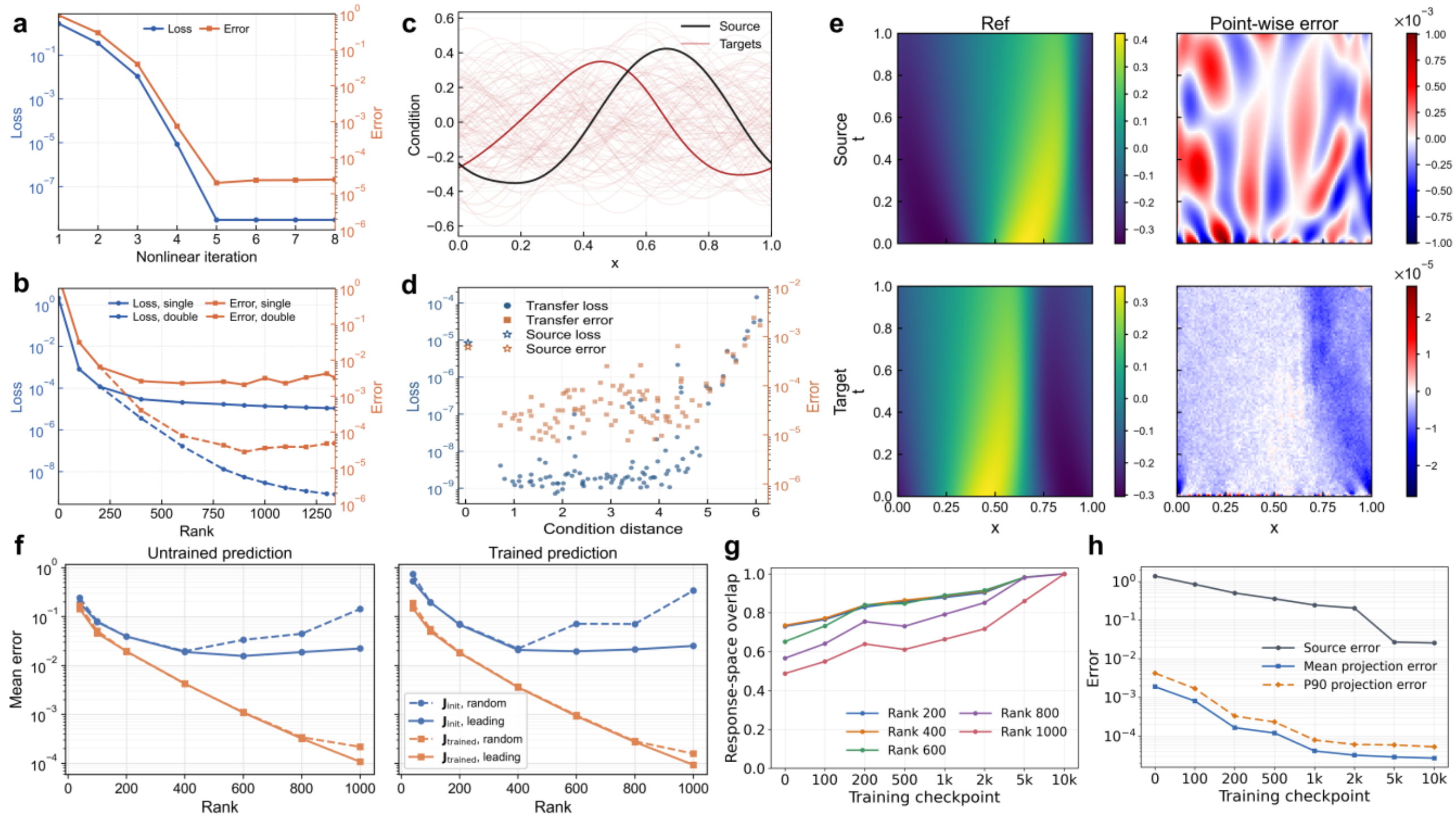

**Fig. 2 | Trained Jacobian response spaces enable accurate cross-condition transfer. a**, Convergence of the response-space nonlinear least-squares solve for the representative target in **c**. **b**, Transfer loss and error versus response-space rank in single and double precision. **c**, Source initial condition (black), target family (light red), and representative target used in **a**, **b** and **e** (dark red). **d**, Transfer loss and error across targets versus condition distance; source-model loss and error are shown for reference. **e**, Reference solutions (left) and signed pointwise errors (right) for the source (top) and representative target (bottom). **f**, Disentangling the contributions of the base prediction and the Jacobian-induced response space. The affine offset is provided by either the untrained network prediction (left) or the trained source prediction (right). For each offset, response modes are constructed from either the untrained or trained output Jacobian using its leading right-singular directions or rank-matched random parameter directions. **g**, Overlap between response spaces at intermediate Adam checkpoints and the final response space at selected ranks. **h**, Evolution of source-model error and mean and 90th-percentile projection errors during Adam training; projection errors use $r = 1000$. All scalar error metrics are relative $L_2$ errors.

For the representative target in Fig. 2c, LST converges within five iterations (0.46 s), reducing both the transfer loss and solution error by several orders of magnitude (Fig. 2a). Full-parameter optimization instead requires thousands of target-specific updates regardless of whether it is initialized randomly or from the trained source model (Supplementary Fig. S4). Generic iterative optimizers applied to the same fixed affine response space are also substantially slower (Supplementary Fig. S5). These controls show that rapid LST adaptation arises from exploiting the least-squares structure of the affine response representation rather than from warm starting or response-space restriction alone.

Increasing the response-space rank improves transfer accuracy (Fig. 2b). At higher ranks, single-precision calculations saturate whereas double precision continues to reduce loss and error, consistent with finite-precision limitations in increasingly ill-conditioned response-coordinate problems. We therefore use double precision and $r$=1000 for subsequent Burgers experiments. At this rank, the representative transferred field closely matches the reference solution with small pointwise errors (Fig. 2e). Across the target family, LST maintains low loss and error over a substantial range of initial-condition distances, but both increase for more distant targets (Fig. 2c,d), revealing the finite effective range of a single source response space. For this Burgers family, post-transfer loss rises alongside solution error beyond this range, providing a practical, label-free indicator of degraded transfer coverage and motivating residual-guided enrichment.

To identify the origin of this transfer capacity, we independently vary the affine offset and the Jacobian used to construct the response modes (Fig. 2f). At high rank, modes derived from the trained Jacobian reduce mean target error by approximately two orders of magnitude regardless of whether the offset is trained or untrained, whereas changing the offset alone has a much smaller effect. The trained Jacobian is therefore the primary source of the large training-associated improvement observed for Burgers. During Adam training, intermediate response spaces progressively approach the final space while their projection errors decrease (Fig. 2g,h), showing that training strengthens a non-trivial response capacity already present at initialization. This behavior is also robust across source-training optimizers and network widths, and source models with errors spanning several orders of magnitude yield comparable typical target accuracies under double-precision LST (Supplementary Figs. S1–S3 and Table S2).

**Residual-guided enrichment of transfer coverage**

The preceding results show that a single source response space has finite effective coverage, motivating ATM to test whether additional source-specific spaces can repair high-loss regions. ATM constructs its source library and routing map from 100 candidate conditions, while an independent unseen test set is used only for evaluation. Acquisition and stopping depend exclusively on candidate transfer losses. In the two-dimensional POD projection, residual-guided acquisition places new sources in initially high-loss regions, progressively reducing transfer loss and contracting the independently evaluated error landscape (Fig. 3a). By stage 5, low loss and error extend across most of the sampled condition domain.

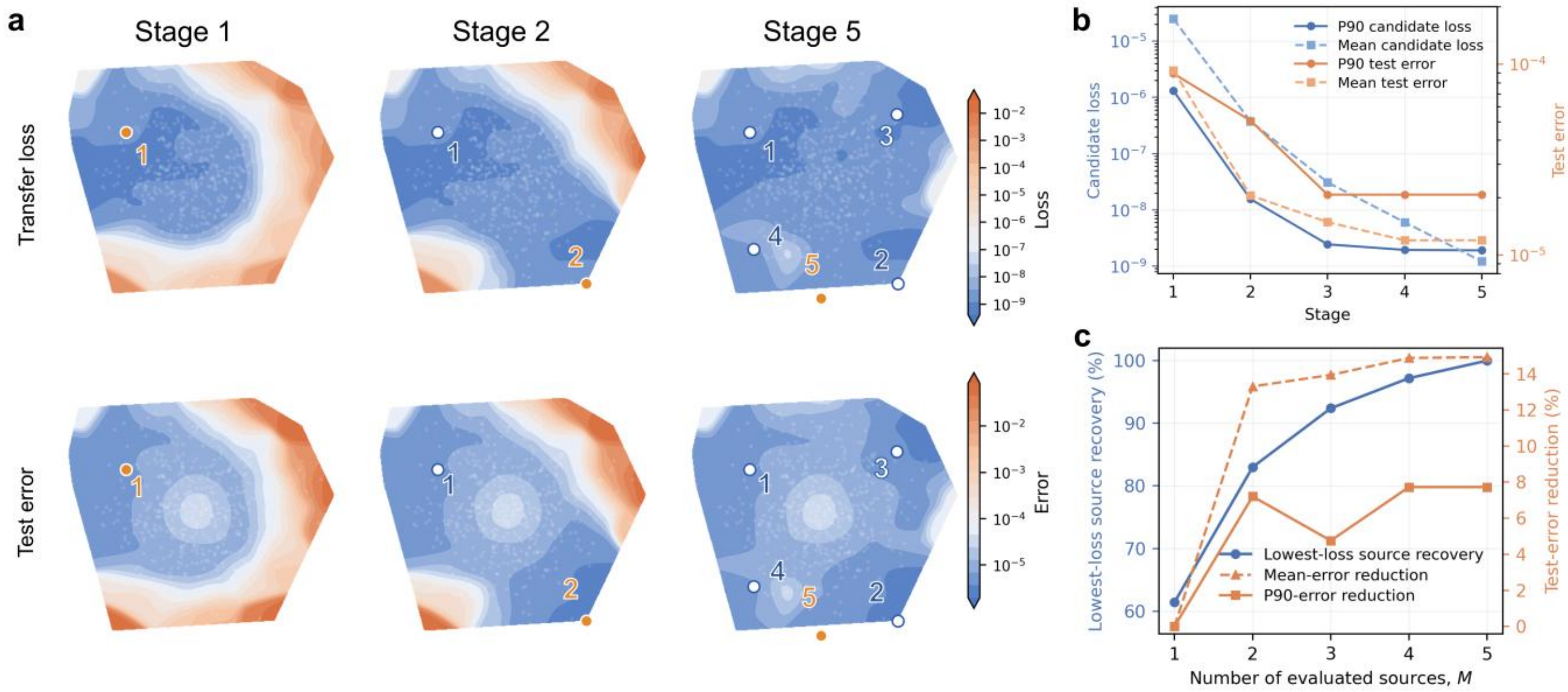

**Fig. 3 | Residual-guided source enrichment progressively expands transfer coverage. a**, Transfer-loss landscapes over the candidate set (top) and solution-error landscapes over an independent test set (bottom) at ATM stages 1, 2 and 5. Initial conditions are represented by their first two proper orthogonal decomposition (POD) coordinates. Grey points denote the evaluated conditions, and numbered markers indicate the source-acquisition order. **b**, Mean and 90th-percentile candidate losses and test errors across enrichment stages. **c**, Post hoc deployment cost–accuracy analysis. For each test condition, sources are ranked by their transfer losses at the nearest candidate condition, and the top $M$ sources are evaluated using LST; $M = 1$ corresponds to the default ATM deployment. Lowest-loss source recovery is the fraction of test conditions for which these $M$ sources include the source yielding the lowest transfer loss among all five acquired sources. Mean and 90th-percentile error reductions are measured relative to $M = 1$.

Across five enrichment stages, the mean and 90th-percentile candidate losses decrease by three to four orders of magnitude, while test errors fall towards $10^{-5}$ (Fig. 3b). Most solution-level improvement occurs by stage 3; later stages further reduce candidate loss but yield only marginal error gains, motivating the 90th-percentile loss reduction as a reference-free stopping criterion. In five matched Burgers constructions, residual-guided acquisition also produces a larger reduction from the initial-stage error than random acquisition (8.27-fold versus 4.21-fold) and a lower final mean error, $(3.18\pm1.50)\times10^{-5}$ versus $(9.71\pm6.80)\times10^{-5}$.

We further examine whether evaluating multiple candidate sources at deployment improves accuracy (Fig. 3c). Increasing the number of evaluated sources from $M = 1$ to $M = 3$ raises recovery of the globally lowest-loss source from approximately 61% to 92%, but reduces mean and 90th-percentile errors by less than about 15% and 8%, respectively. Because the lowest-loss source need not minimize reference error, these gains are not strictly monotonic. The default single-source routing strategy therefore retains most of the observed accuracy with one LST solve.

## Accuracy–cost comparison with global operators

To characterize the accuracy–cost trade-off between local response-space construction and global operator learning, we compare ATM with data-driven and physics-informed DeepONets on the same Burgers initial-condition family. The two DeepONet baselines share the same architecture and are trained over 1,000 conditions, using labeled condition–solution pairs and PDE-system residuals, respectively. ATM instead reuses the five-source library constructed above. All methods are evaluated on the same 200 unseen test conditions. Detailed network, training and evaluation settings are provided in Supplementary Section S3.2. We report empirical solution accuracy together with measured construction and per-target deployment costs under each method's native modeling protocol.

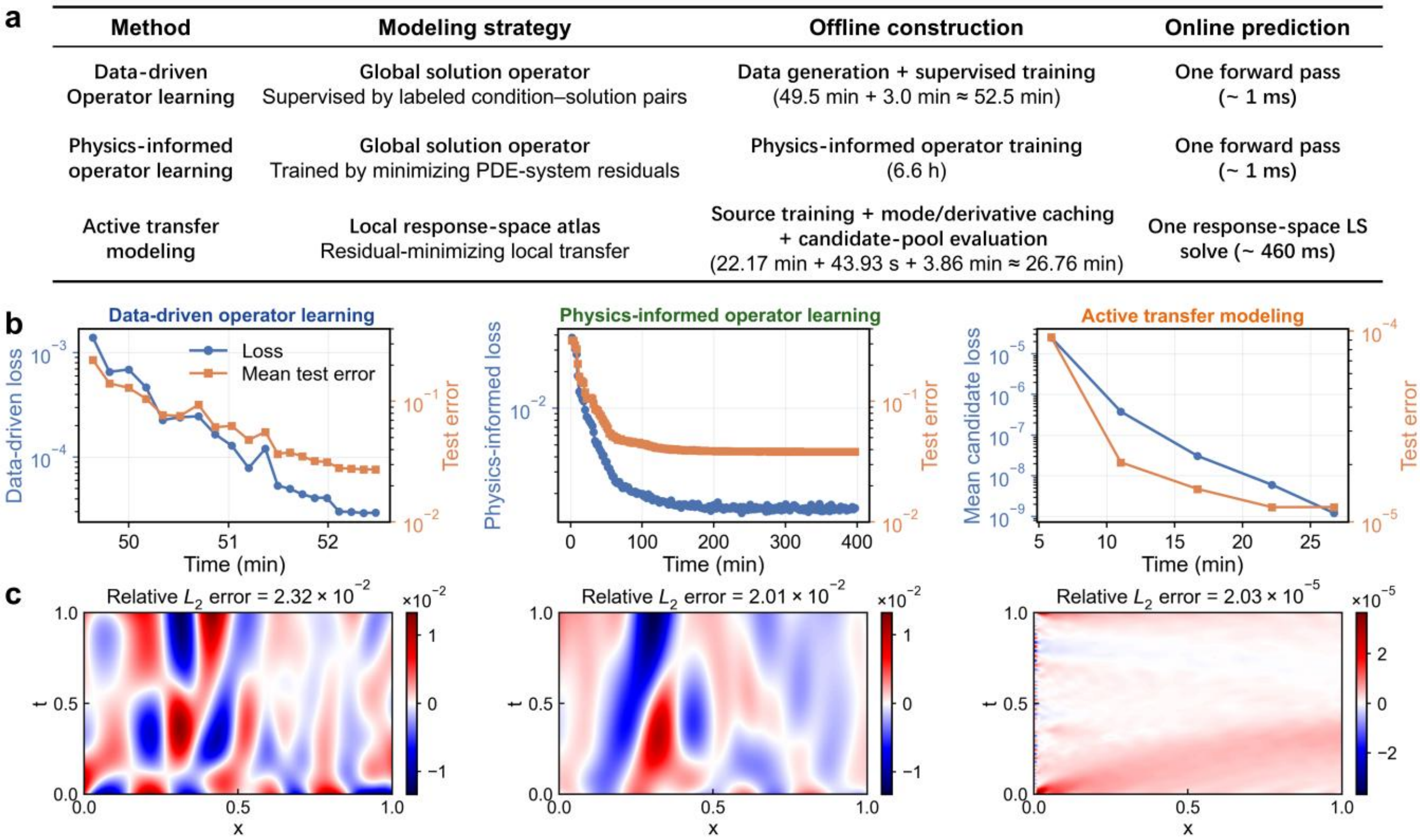

| Method | Modeling strategy | Offline construction | Online prediction |
|---|---|---|---|
| Data-driven Operator learning | Global solution operator Supervised by labeled condition–solution pairs | Data generation + supervised training (49.5 min + 3.0 min ≈ 52.5 min) | One forward pass (~ 1 ms) |
| Physics-informed operator learning | Global solution operator Trained by minimizing PDE-system residuals | Physics-informed operator training (6.6 h) | One forward pass (~ 1 ms) |
| Active transfer modeling | Local response-space atlas Residual-minimizing local transfer | Source training + mode/derivative caching + candidate-pool evaluation (22.17 min + 43.93 s + 3.86 min ≈ 26.76 min) | One response-space LS solve (~ 460 ms) |

**Fig. 4 | Accuracy–cost trade-offs between ATM and global operator baselines. a**, Comparison of the modeling strategies and measured offline construction and per-target online costs of data-driven DeepONet, physics-informed DeepONet and ATM on the Burgers benchmark. ATM construction includes training five source models, caching their response modes and PDE-relevant derivatives, and evaluating the candidate set to construct the routing map. **b**, Method-specific construction histories. For the two DeepONet baselines, the training objective and mean test error are shown over optimization; for ATM, the mean candidate transfer loss and mean test error are shown across enrichment stages. Test errors are evaluated on the same 200 unseen conditions and are used only for post hoc assessment. **c**, Signed pointwise error fields for a representative unseen target. Titles report the errors, and per-target online costs are summarized in **a**. Color limits are independently scaled and centered at zero to reveal the spatial structure of each error field.

Because the three methods optimize different objectives, their loss values should not be compared directly (Fig. 4b). For both DeepONet baselines, decreasing training objectives are accompanied by reductions in mean test error, which reaches $2.72\times10^{-2}$ for the data-driven model and $3.83\times10^{-2}$ for the physics-informed model. ATM follows a different construction trajectory: successive enrichment stages add source-specific response spaces, progressively reducing the mean candidate transfer loss and yielding a mean test error of $1.20\times10^{-5}$ after five stages. Thus, the global operators refine a single shared mapping through continued parameter optimization, whereas ATM extends parametric coverage by incrementally adding local response spaces. The representative unseen target in Fig. 4c shows the same ordering in solution error.

The data-driven and physics-informed operators require 52.5 min and 6.6 h of offline construction, respectively, compared with 26.76 min for the five-stage ATM library. An L-BFGS sensitivity test reduces the physics-informed DeepONet error to $1.12\times10^{-2}$ but requires 8 h (Supplementary Fig. S6). Notably, the initial single-source ATM stage already attains a mean error approximately two orders of magnitude below the fully trained operator baselines after about 5 min of construction. The trade-off is deployment speed: the global operators require approximately 1 ms per target, whereas ATM performs one routed LST solve in approximately 460 ms. On this Burgers benchmark, ATM therefore exchanges one-shot inference for subsecond target-specific adaptation while allowing parametric coverage to be extended incrementally.

**Transferability and enrichment across parametric systems**

We next test whether single-condition response-space transfer and active enrichment extend beyond Burgers across five additional benchmark systems spanning function-valued forcing and source terms, spatially varying coefficients and permeability fields, and nonlinear initial-condition variation (Fig. 5a). Unlike the other systems, Navier–Stokes uses a time-marching LST variant with fixed spatial response modes derived from the source model and time-varying response coordinates (Methods and Supplementary Section S2.4). Table 1 summarizes source-model, single-source LST and final ATM accuracies together with construction and online costs. All physics-informed operator baselines were trained and evaluated on the same computing platform and under the same benchmark definitions as ATM. The four benchmarks overlapping with Wang et al.[7] follow the corresponding published configurations. For Darcy and Navier–Stokes, we retain PINO architectures following the original study but retrain them on our benchmark definitions, which differ from the original PINO benchmarks in the coefficient-field sampling and initial-vorticity initialization, respectively (Methods and Supplementary Section S3.2).

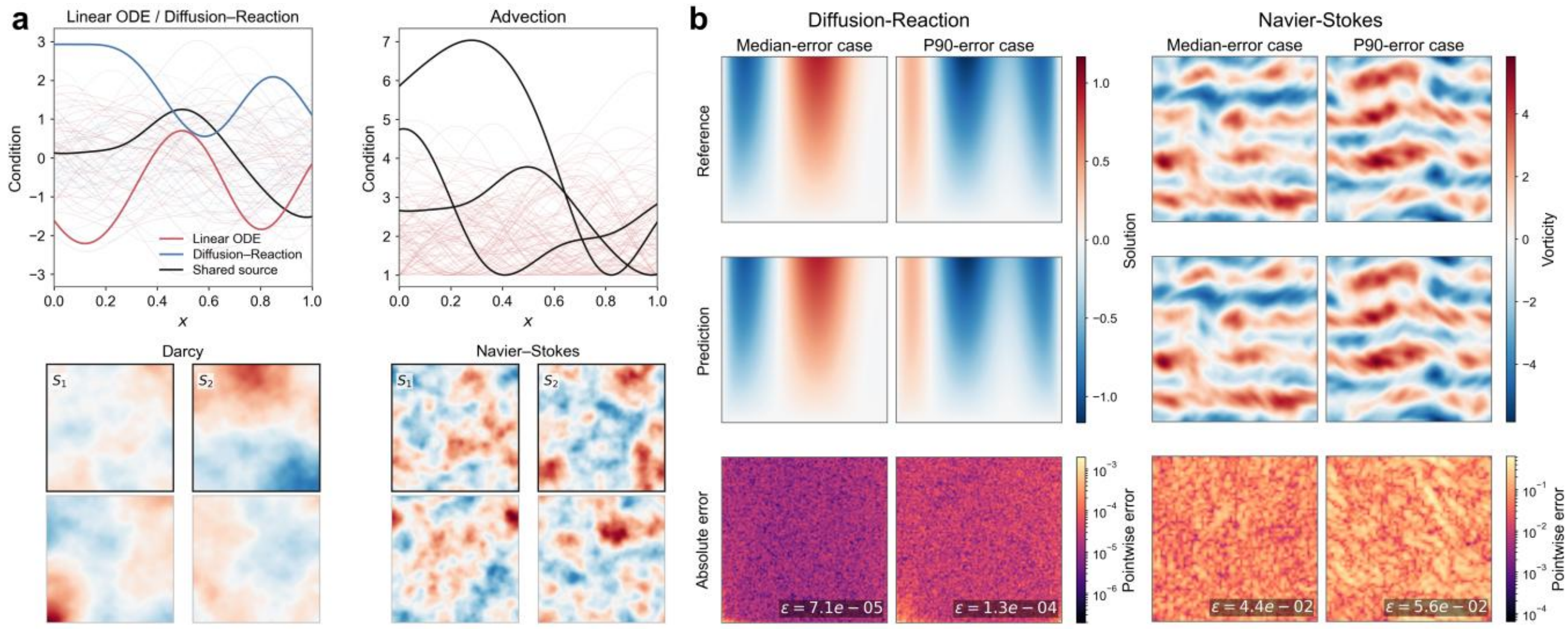

**Fig. 5 | Response-space transfer and enrichment across diverse parametric systems. a**, Representative condition families for the linear ODE, diffusion–reaction, advection, Darcy and Navier–Stokes benchmarks. Curves denote forcing, source and transport-coefficient functions, whereas fields denote permeability and initial-vorticity conditions. Faint curves and unlabeled fields show representative sampled conditions; highlighted curves and fields labeled $s_i$ indicate acquired sources. **b**, Reference solutions, ATM predictions and absolute pointwise errors for test cases near the median and 90th percentile of the final error distribution for diffusion–reaction and Navier–Stokes. Navier–Stokes fields show vorticity at final time, $t = 0.5$, obtained using time-marching LST. Reported values are relative $L_2$ errors.

Single-source LST supports cross-condition transfer across all six systems. In four benchmarks, the mean error of single-source LST on unseen targets is lower than the mean error of the trained single-condition source models at their own training conditions (Table 1), showing that cross-condition recovery is not simply limited by source-prediction accuracy. By combining the source-derived response representation with the target governing equations, LST can attain aggregate accuracies exceeding those of direct single-condition neural solves. Active enrichment, however, is strongly problem dependent: Burgers, Darcy and diffusion–reaction improve substantially, whereas the linear ODE is already saturated and standard advection and Navier–Stokes show little or no solution-level gain. Expanded-range advection confirms that the negligible gain in the standard family reflects broad single-source coverage: when larger transport-time variation creates high-loss regions, additional sources again improve target accuracy (Supplementary Section S4.3). Thus, transferability does not imply enrichment benefit; additional sources are most useful when they provide complementary representation capacity for poorly represented targets.

**Table 1 | Accuracy and computational costs of ATM and global physics-informed operators across six parametric systems.** Errors are relative $L_2$ errors and are reported as mean ± s.d. where applicable. Source-model errors are evaluated across the retained source conditions, with the final source-library size shown in parentheses. ATM stage-1 and final errors are evaluated on unseen test conditions before and after active enrichment, respectively.

Global PI-operator offline time denotes operator training time, whereas ATM offline time includes source-model training, response-space and cache construction, and candidate-set evaluation throughout active enrichment. All offline times were measured in our runs on the same hardware. ATM online time is reported per unseen target using the final source library and default single-source routing; it corresponds to one LST solve for the first five systems and to the complete sequence of modal-coordinate updates for Navier–Stokes. The first four global PI-operator baselines use PI-DeepONet following Wang et al.[7]. †For Darcy and Navier–Stokes, we use PINO baselines following Li et al.[8], retrained and evaluated under the same benchmark definitions as ATM. These are treated as reference comparisons rather than directly matched counterparts to the other four operator baselines because the PINO architecture, differentiation scheme and residual implementation differ; moreover, our Darcy and Navier–Stokes benchmark definitions differ from the original PINO benchmarks in coefficient-field sampling and initial-condition generation, respectively.

| System | Condition | Global PI operator | | ATM | | | | |
|---|---|---|---|---|---|---|---|---|
| | | Error | Offline | Source error (n) | Stage-1 error | Final error | Offline | Online |
| Linear ODE | Forcing terms | $2.91\times10^{-3} \pm 2.39\times10^{-3}$ | 2.92 min | $1.54\times10^{-4} \pm 9.18\times10^{-5}$ (2) | $\mathbf{9.68\times10^{-8} \pm 7.14\times10^{-8}}$ | $\mathbf{9.73\times10^{-8} \pm 7.20\times10^{-8}}$ | 0.2 min | 0.0019 s |
| Diffusion-reaction | Source terms | $5.14\times10^{-3} \pm 3.11\times10^{-3}$ | 13.86 min | $1.79\times10^{-3} \pm 4.96\times10^{-4}$ (3) | $\mathbf{1.09\times10^{-4} \pm 6.16\times10^{-5}}$ | $\mathbf{8.22\times10^{-5} \pm 4.40\times10^{-5}}$ | 8.12 min | 0.45 s |
| Burgers | Initial conditions | $3.83\times10^{-2} \pm 4.17\times10^{-2}$ | 6.6 h | $4.72\times10^{-4} \pm 2.18\times10^{-4}$ (5) | $\mathbf{9.29\times10^{-5} \pm 4.78\times10^{-4}}$ | $\mathbf{1.20\times10^{-5} \pm 1.06\times10^{-5}}$ | 26.76 min | 0.46 s |
| Advection | Variable coefficients | $3.91\times10^{-2} \pm 1.12\times10^{-2}$ | 51.87 min | $2.58\times10^{-3} \pm 8.71\times10^{-4}$ (3) | $\mathbf{4.56\times10^{-3} \pm 2.09\times10^{-3}}$ | $\mathbf{4.57\times10^{-3} \pm 2.10\times10^{-3}}$ | 3.09 min | 0.10 s |
| Darcy | Permeability fields | $2.61\times10^{-3} \pm 1.58\times10^{-3\dagger}$ | 32.63 min | $7.04\times10^{-3} \pm 1.87\times10^{-3}$ (3) | $\mathbf{6.50\times10^{-4} \pm 3.54\times10^{-4}}$ | $\mathbf{3.93\times10^{-4} \pm 1.15\times10^{-4}}$ | 17.26 min | 0.39 s |
| Navier-Stokes | Initial conditions | $6.34\times10^{-2} \pm 3.40\times10^{-3\dagger}$ | 7.0 h | $3.85\times10^{-2} \pm 2.03\times10^{-4}$ (2) | $\mathbf{4.38\times10^{-2} \pm 7.67\times10^{-3}}$ | $\mathbf{4.52\times10^{-2} \pm 7.80\times10^{-3}}$ | 44.85 min | 1.97 s |

Representative diffusion–reaction and Navier–Stokes cases illustrate this range of behavior (Fig. 5b). Diffusion–reaction reaches errors on the order of $10^{-4}$, including for a case near the 90th percentile of the final error distribution, whereas time-marching LST for Navier–Stokes preserves the dominant vorticity structures with errors on the order of $10^{-2}$. Per-target adaptation spans milliseconds to seconds, with Navier–Stokes incurring the largest cost because its response coordinates evolve throughout the trajectory. These results show that the same source-response principle extends across qualitatively different differential-equation systems, while the attainable accuracy and adaptation cost remain problem dependent. Across the evaluated physics-informed operator comparisons, final ATM models achieve lower mean errors and lower offline construction costs while retaining target-specific online adaptation.

As a mechanistic control, we evaluate matched initialization-state response representations for three independently sampled sources per benchmark. All 18 trained representations yield lower mean target errors, while the initialization states retain non-trivial recovery capacity, indicating that training strengthens rather than creates transferable response structure from scratch (Supplementary Table S6). As a single-source trajectory control for Navier–Stokes, a POD space retaining virtually all source-trajectory energy nevertheless yields a mean projection error of 0.87 on unseen initial conditions, showing that accurate compression of one trajectory does not by itself provide the off-trajectory representation captured by the Jacobian-derived response space.

## Discussion

Our results show that a neural solution model trained at a single condition can encode substantially more than a condition-specific approximation. Its output Jacobian defines a field-space response representation that can capture solution variations induced by changes in forcing, coefficients or initial conditions, allowing the model to function as a reusable local parametric model. This interpretation does not require the dominant Jacobian modes to coincide with physical parameter sensitivities; rather, their span provides an approximation space in which LST recovers target solutions by minimizing the target PDE-system residual. The consistency of this behavior across independently trained sources and different systems suggests that cross-condition transfer is not specific to a particular source model. Initialization-state controls further indicate that training strengthens a non-trivial response representation already present at initialization rather than creating transferable capacity entirely from scratch.

LST makes this response structure operational without adapting the original network weights. The target solution is sought in a fixed affine field space while retaining the full target PDE-system residual, yielding a structured least-squares problem over response coordinates. Its rapid convergence cannot be explained simply by source initialization or fewer optimization variables: standard iterative optimizers applied to the same rank-1,000 affine response model do not reproduce the few-step convergence of LST. Instead, LST explicitly exploits the least-squares structure of the fixed response representation, while precomputing response modes and their PDE-relevant derivatives avoids repeated network evaluation and differentiation during target adaptation. Structure-exploiting coordinate updates and reusable PDE-ready response information therefore provide complementary sources of computational efficiency and distinguish LST from conventional target-stage fine-tuning.

This local-response perspective also clarifies the relation to global operator learning. Global operators construct coverage over a prescribed condition domain during offline training and amortize this cost through rapid inference. LST and ATM instead build from local source models, retain a target-specific physics solve and extend coverage incrementally. Across the evaluated comparisons, this approach achieved lower errors and lower offline construction costs than the physics-informed operator baselines, while requiring millisecond-to-second target adaptation rather than one-shot inference. The experiments also show that transferability alone does not determine the benefit of enrichment: a single source may already provide broad coverage, whereas additional sources are most useful when they provide complementary representation capacity for poorly represented targets. Accordingly, the post-transfer residual used by ATM should be regarded as a practical coverage indicator rather than a rigorous estimator of solution error.

Several limitations remain. Each source provides only a rank-truncated affine approximation to a generally nonlinear solution family, with no general guarantee that this local representation remains accurate under finite or strongly nonlinear condition changes, and increasing rank eventually encounters conditioning and finite-precision limitations. Cached response fields and derivatives also increase memory requirements with state dimension and response-space rank, while candidate-based acquisition and nearest-candidate routing may require larger candidate sets or more informative condition metrics in higher-dimensional or more complex condition spaces. More broadly, the present experiments focus on physics-informed neural solution models and a limited range of architectures and condition families; how transferable response structure depends on architecture, training objective, geometric variation and stronger out-of-distribution shifts remains to be established. These limitations notwithstanding, our results support a broader view of neural PDE solvers: a model trained at one condition can serve as a reusable local parametric model whose response structure can be queried and extended on demand.

# Methods

## Source response-space construction

Consider a parametric differential-equation system $f(q(\boldsymbol{x},t);\mu)=0, \mu\in\mathcal{P}$, where $q$ denotes a scalar- or vector-valued solution field and $\mathcal{P}$ is the condition space. Depending on the problem, $\mu$ may represent an initial or boundary condition, a forcing or source term, or a spatially varying coefficient field. For a selected source condition $\mu^{(j)}$, we train a neural solution model $q(\boldsymbol{x},t;\boldsymbol{\theta}_0)$ by minimizing a physics-informed loss comprising the governing-equation residual and the applicable initial- and boundary-condition terms. The trained parameters and corresponding source prediction are denoted by $\boldsymbol{\theta}_0^{(j)}$ and $q_0^{(j)}(\boldsymbol{x},t)=q(\boldsymbol{x},t;\boldsymbol{\theta}_0^{(j)})$, respectively. The resulting source model, together with its cached response-space representation, is denoted by $s_j$. In the following single-source construction, we consider a fixed source and suppress the superscript $(j)$ for notational clarity.

The pointwise output Jacobian of the trained source model is

$$\mathbf{J}(\boldsymbol{x},t)=\left.\frac{\partial q(\boldsymbol{x},t;\boldsymbol{\theta})}{\partial\boldsymbol{\theta}}\right|_{\theta=\theta_0} \quad (1)$$

The collection of pointwise Jacobians defines a linear map from parameter perturbations to output-field perturbations. The range of this field-valued Jacobian operator—represented numerically by the stacked sampled output Jacobian—defines the local output tangent space of the neural solution model. Following the matrix-free randomized construction introduced in LSR[37], we approximate the dominant right-singular subspace of the sampled output Jacobian without explicitly forming the full matrix. Let $\mathbf{V} = [\boldsymbol{v}_1, \boldsymbol{v}_2, \ldots, \boldsymbol{v}_r] \in \mathbb{R}^{m \times r}$ denote the retained rank-$r$ parameter basis, where $m$ is the number of trainable parameters. The corresponding solution-field response modes are $\phi_j(\boldsymbol{x},t) = \mathbf{J}(\boldsymbol{x},t)\boldsymbol{v}_j$, or, collectively, $\mathbf{\Phi}(\boldsymbol{x},t) = \mathbf{J}(\boldsymbol{x},t)\mathbf{V}$. These modes define the rank-truncated affine response space

$$\mathcal{A} = q_0 + \mathrm{span}(\mathbf{\Phi}) = \{q_0(\boldsymbol{x},t) + \mathbf{\Phi}(\boldsymbol{x},t)\boldsymbol{\alpha} : \boldsymbol{\alpha} \in \mathbb{R}^r\}. \tag{2}$$

Although the directions $\mathbf{V}$ are represented in parameter coordinates, the transferred representation is constructed and optimized in output-field space through $\mathbf{\Phi} = \mathbf{JV}$. Because a source model may be reused for many target conditions, we precompute and cache $q_0$, the response modes $\mathbf{\Phi}$, and all spatial and temporal derivatives required by the governing PDE system. These quantities are evaluated by explicit differentiation during source-cache construction. Subsequent target residuals can therefore be assembled from the cached fields and derivatives without repeatedly evaluating or differentiating the source network.

**Output-tangent interpretation**

The affine response space in Eq. (2) admits a local interpretation in relation to the parametric solution family. Let $q^*(\mu)$ denote the exact solution, and consider a differentiable condition path $\mu(\varepsilon)$ passing through the fixed source condition at $\varepsilon = 0$. Suppose that, locally along this path, the solution can be represented by a differentiable path of network parameters $\boldsymbol{\theta}^*(\varepsilon)$, up to a neural representation error $e(\varepsilon)$:

$$q^*(\mu(\varepsilon)) = q(\cdot;\boldsymbol{\theta}^*(\varepsilon)) + e(\varepsilon), \boldsymbol{\theta}^*(0) = \boldsymbol{\theta}_0 \tag{3}$$

Differentiating at the source condition gives

$$\left.\frac{\mathrm{d}q^*(\mu(\varepsilon))}{\mathrm{d}\varepsilon}\right|_{\varepsilon=0} = J\left.\frac{\mathrm{d}\boldsymbol{\theta}^*(\varepsilon)}{\mathrm{d}\varepsilon}\right|_{\varepsilon=0} + \left.\frac{\mathrm{d}e(\varepsilon)}{\mathrm{d}\varepsilon}\right|_{\varepsilon=0} \tag{4}$$

Thus, up to variations in the neural representation error, first-order solution changes induced by condition perturbations can be represented within the range of the source-model output Jacobian. The rank-$r$ space in Eq. (2) replaces this full range by $\mathrm{span}(\mathbf{\Phi})$. Its transferability therefore depends on how well these retained field directions approximate condition-induced variations of the solution family. This interpretation does not require the dominant Jacobian modes to coincide with physical parameter sensitivities. For finite changes in condition, transfer accuracy further depends on how well the truncated affine response space approximates the generally nonlinear parametric solution family away from the source condition.

**Linearized Subspace Transfer**

For a target condition $\mu$ and a fixed source model $s$, LST seeks the target solution in the source-derived affine response space, $q_{\mathrm{LST}}(\boldsymbol{x},t;\mu,s,\boldsymbol{\alpha})=q_0(\boldsymbol{x},t)+\boldsymbol{\Phi}(\boldsymbol{x},t)\boldsymbol{\alpha}$, where the dependence of $q_0$ and $\boldsymbol{\Phi}$ on the source is suppressed for notational clarity. The source-network parameters remain fixed throughout target adaptation. Importantly, LST optimizes the affine output representation $q_0+\boldsymbol{\Phi\alpha}$; it does not evaluate or optimize the nonlinear network $q(\boldsymbol{x},t;\boldsymbol{\theta}_0+\mathbf{V}\boldsymbol{\alpha})$. An LST-transferred field therefore need not correspond to the output of the original network at any actual parameter state.

Let $\boldsymbol{f}(\boldsymbol{\alpha};\mu,s)\in\mathbb{R}^{N_0}$ denote the discretized target PDE-system residual obtained by evaluating and stacking the governing-equation residual and the applicable initial- and boundary-condition residuals at their respective sampling points. Here $N_0$ is the total number of residual components. The target-specific response coordinates are determined by

$$\boldsymbol{\alpha}^*(\mu;s)=\arg\min_{\boldsymbol{\alpha}}\frac{1}{N_0}\left\|\boldsymbol{f}(\boldsymbol{\alpha};\mu,s)\right\|_2^2 \tag{5}$$

The field $q_{\mathrm{LST}}$ and all of its PDE-relevant spatial and temporal derivatives depend affinely on $\boldsymbol{\alpha}$. The target residual itself, however, generally remains nonlinear in α when the governing equations are nonlinear. LST therefore solves a structured nonlinear least-squares problem over a fixed, source-derived affine field representation while retaining the full target PDE-system residual. This differs from LSR, which linearizes the residual around the trained parameter state and solves the resulting linear least-squares problem for same-instance refinement; LST linearizes the neural output representation but not the governing residual, and uses the resulting response space for cross-condition transfer.

Because the response representation is fixed, the quantities required to evaluate the residual and form updates with respect to α can be assembled directly from the cached source prediction, response modes and PDE-relevant derivatives. LST can therefore exploit the least-squares structure in response coordinates without differentiating through or re-optimizing the source network during target adaptation. This structure-exploiting coordinate solve reduces the number of target-stage iterations, whereas the PDE-ready cache reduces the computational cost of each residual evaluation; their respective contributions are examined separately in the Supplementary Information.

After convergence, the transferred solution is

$$q_{\mathrm{LST}}(\boldsymbol{x},t;\mu,s)=q_0(\boldsymbol{x},t)+\boldsymbol{\Phi}(\boldsymbol{x},t)\boldsymbol{\alpha}^*(\mu;s) \tag{6}$$

and the associated transfer loss is defined as

$$\ell(\mu;s)=\frac{1}{N_0}\left\|\boldsymbol{f}(\boldsymbol{\alpha}^*(\mu;s);\mu,s)\right\|_2^2 \tag{7}$$

The accuracy attained by LST reflects both the approximation capacity of the truncated affine response space and the conditioning and convergence of the residual-coordinate solve. Increasing the response-space rank can enlarge the available representation, but weak or poorly conditioned response directions need not improve the numerically recovered solution.

**Active Transfer Modeling**

Active Transfer Modeling (ATM) extends single-source LST by maintaining an incrementally enriched library of source-specific response spaces together with a loss-based routing map defined over a finite candidate set, $\mathcal{P}_{\text{cand}} = \{\mu_i\}_{i=1}^{N_c}$. The initial source condition $\mu^{(1)}$ is sampled independently from the same underlying distribution, whereas all subsequent source conditions are selected using the residual-guided acquisition criterion described below. Let $\mathcal{S}_k = \{s_1, \dots, s_k\}$ denote the available source library after $k$ sources have been constructed. For each candidate condition $\mu_i$, ATM records the lowest transfer loss achieved by the current source set, $\ell_k(\mu_i) = \min_{s \in \mathcal{S}_k} \ell(\mu_i; s)$, together with the corresponding loss-optimal source assignment, $a_k(\mu_i) = \arg\min_{s \in \mathcal{S}_k} \ell(\mu_i; s)$. Because $\ell_k(\mu_i)$ is computed from the post-transfer PDE-system residual, construction of this routing map requires no reference solutions at the candidate conditions. We use the incumbent transfer loss $\ell_k$ as a practical indicator of transfer coverage rather than as an estimator of solution error, because the correspondence between residual and solution error is problem dependent.

The next source condition is selected at the candidate with the largest current transfer loss,

$$\mu^{(k+1)} = \arg\max_{\mu_i \in \mathcal{P}_{\text{cand}}} \ell_k(\mu_i) \tag{8}$$

After training the new source model $s_{k+1}$ at $\mu^{(k+1)}$ and constructing its response-space cache, LST is evaluated from $s_{k+1}$ to every candidate condition, yielding the new-source transfer losses $\ell(\mu_i; s_{k+1})$. The incumbent transfer loss is then updated pointwise as $\ell_{k+1}(\mu_i) = \min\{\ell_k(\mu_i), \ell(\mu_i; s_{k+1})\}$ with the corresponding source assignment

$$a_{k+1}(\mu_i) = \begin{cases} s_{k+1}, & \ell(\mu_i; s_{k+1}) < \ell_k(\mu_i) \\ a_k(\mu_i), & \text{otherwise} \end{cases} \tag{9}$$

Consequently, $\ell_{k+1}(\mu_i) \le \ell_k(\mu_i)$ for every candidate condition. Each enrichment stage therefore preserves the lowest transfer loss previously attained at every candidate while requiring new transfer evaluations only from the newly acquired source; losses and assignments associated with the existing sources are retained in the cache.

To determine when further enrichment is no longer warranted, ATM monitors the 90th percentile of the incumbent candidate-loss distribution, $L_k^{90} = \mathrm{P}90\{\ell_k(\mu_i)\}_{i=1}^{N_c}$. Enrichment terminates when the relative reduction between two consecutive stages, $\rho_k = (L_{k-1}^{90} - L_k^{90}) / L_{k-1}^{90}$, falls below a prescribed threshold, taken as 10% in this work. The percentile-based criterion is less sensitive than the maximum loss to isolated high-loss candidates while remaining responsive to changes in the upper tail of the transfer-loss distribution.

For an unseen target condition $\mu$, ATM identifies the nearest candidate,

$$i^* = \arg\min_{1\le i\le N_c} d(\mu, \mu_i) \tag{10}$$

selects the cached source assignment $s^* = a_K(\mu_{i^*})$ and performs one LST solve at the target condition using $s^*$. Here, $d(\mu, \mu_i)$ denotes the Euclidean distance between the discretized condition representations. The nearest-candidate lookup therefore determines only which source response space is used.

**Time-marching LST for Navier–Stokes dynamics**

For the Navier–Stokes benchmark, we use a time-marching variant of LST to accommodate the evolving response coordinates required over a long nonlinear trajectory. Unlike the target-level space–time representations used for the other systems, a fixed spatial response basis is constructed from the source-model output Jacobian at the initial time, $\mathbf{\Phi}(\boldsymbol{x}) = \mathbf{J}(\boldsymbol{x}, 0)\mathbf{V}$, and reused throughout the trajectory. At time $t_n$, the transferred vorticity field is represented as $q_{\mathrm{LST}}(\boldsymbol{x}, t_n) = q_0(\boldsymbol{x}, t_n) + \mathbf{\Phi}(\boldsymbol{x})\boldsymbol{\alpha}_n$, where a separate response-coordinate vector $\boldsymbol{\alpha}_n$ is determined at each time step. Thus, the source-derived response basis remains fixed while the target-specific coordinates evolve in time. The initial coordinates are obtained from a regularized projection of the target initial condition, and the subsequent coordinates are computed sequentially using damped least-squares updates that enforce the time-discretized governing residual. This construction retains a single source-derived spatial response space while allowing the transferred trajectory to move within that space over time. Details of the temporal discretization, regularization, damping strategy and linear-algebra implementation are provided in Supplementary Section S2.4.

**Evaluation measures**

Solution accuracy is measured using the relative discrete $L_2$ error

$$\varepsilon = \frac{\left\| q_{\mathrm{pred}} - q_{\mathrm{ref}} \right\|_2}{\left\| q_{\mathrm{ref}} \right\|_2} \tag{11}$$

evaluated on the benchmark-specific reference grid. For vector-valued fields, the norm includes all reported field components. To separate response-space representation capacity from the performance of the residual-based LST solve, we define the affine projection error of a target solution $q_{\mathrm{ref}}(\mu)$ as

$$\varepsilon_{\mathrm{proj}}(\mu; s, r) = \min_{\boldsymbol{\alpha}\in\mathbb{R}^r} \frac{\left\| q_{\mathrm{ref}}(\mu) - q_0^{(s)} - \mathbf{\Phi}_r^{(s)}\boldsymbol{\alpha} \right\|_2}{\left\| q_{\mathrm{ref}}(\mu) \right\|_2} \tag{12}$$

where $\mathbf{\Phi}_r^{(s)}$ contains the first $r$ response modes. This metric measures the best representation available within the affine source response space independently of the residual-based LST solver. It requires the reference solution and is used only for analysis, not during LST or ATM construction.

Response-space similarity is quantified using principal angles. The response modes are evaluated on the same discrete grid, and each rank-$r$ response space is orthonormalized using the standard Euclidean inner product. For two orthonormal basis matrices $\mathbf{Q}_A, \mathbf{Q}_B \in \mathbb{R}^{N_q \times r}$, the response-space overlap is defined as

$$O_r(\mathbf{Q}_A^T, \mathbf{Q}_B) = \frac{1}{r}\left\|\mathbf{Q}_A^T \mathbf{Q}_B\right\|_F^2 = \frac{1}{r}\sum_{j=1}^{r}\cos^2\varphi_j \tag{13}$$

where $\varphi_j$ are their principal angles. The overlap is invariant to rotations within either basis and ranges from zero for mutually orthogonal spaces to one for identical spaces.

## Data availability

The data used in this study are either obtained from previously published works or generated using the numerical procedures described in the paper. All publicly available datasets are cited in the main text. Any additional data required to reproduce the main results are provided together with the source code.

## Code availability

The source code implementing Linearized Subspace Transfer and Active Transfer Modeling, together with the benchmark configurations and scripts used to reproduce the main results, is publicly available at https://github.com/Cao-WenBo/active-transfer-modeling. Additional scripts and intermediate numerical results used for sensitivity and diagnostic analyses are available from the corresponding author upon reasonable request.

## Competing interests

The authors declare no competing interests.

## Acknowledgements

We would like to acknowledge the support of the National Natural Science Foundation of China (No. 92152301).

## Author contributions

W.C.: Conceptualization, Methodology, Software, Formal analysis, Investigation, Visualization, Writing – original draft. W.Z.: Supervision, Methodology, Formal analysis, Writing – review & editing. Both authors discussed the results and approved the final manuscript.

# Supplementary Information

Wenbo Cao, Weiwei Zhang*

## S1. Benchmark definitions and common formulation

### S1.1 Benchmark systems and varying-condition families

We consider six benchmark systems adapted from two established studies on physics-informed operator learning. The first four benchmarks—the linear antiderivative problem, diffusion-reaction equation, viscous Burgers equation, and variable-coefficient advection equation—follow the problem definitions introduced by Wang *et al.*[1]. Together, they form a controlled hierarchy spanning an ordinary differential equation and time-dependent PDEs, linear and nonlinear solution mappings, and functional variability introduced through forcing functions, initial conditions, and spatially distributed coefficients. This diversity allows us to examine whether response spaces extracted from single-condition neural solution models remain transferable across distinct classes of parametric differential equations.

The two-dimensional Darcy and Navier–Stokes systems are based on the governing equations considered by Li et al.[2], but use condition distributions defined for the present study. The steady Darcy problem tests transfer across heterogeneous permeability fields, whereas the Navier–Stokes benchmark considers two-dimensional Kolmogorov flow with randomly varying initial vorticity fields. Its nonlinear advection, incompressibility constraint, multiscale spatiotemporal evolution, and sensitivity to the initial state make it a demanding test of whether a response space extracted from a single-condition neural solution model can support transfer across complex flow dynamics.

The governing equations, initial and boundary conditions, and distributions of the varying conditions are summarized in Table S1. Benchmark-specific network architectures, residual sampling strategies, source-training settings, and LST parameters are reported separately in Sections S2 and S3.

Table S1. Benchmark systems, computational domains and varying conditions. In the first five benchmarks, $q$ denotes the scalar solution field. For the Navier–Stokes system, we retain the standard notation $\boldsymbol{u}=(u,v)$ for velocity and $\omega$ for vorticity.

| System | Governing problem | Domain | Varying condition | Condition generation |
|---|---|---|---|---|
| Linear ODE | $q_x = h(x), q(0)=0$ | $[0,1]$ | Forcing function $h(x)$ | $h \sim \mathcal{GP}(0,K)$ |
| Diffusion-reaction | $q_t - 0.01q_{xx} - 0.01q^2 = h(x)$<br>$q(x,0)=0, q(0,t)=q(1,t)=0$ | $[0,1]\times[0,1]$ | Source function $h(x)$ | $h \sim \mathcal{GP}(0,K)$ |
| Burgers | $q_t + qq_x - 0.01q_{xx} = 0$<br>$q(0,t)=q(1,t), q_x(0,t)=q_x(1,t)$ | $[0,1]\times[0,1]$ | Initial condition $q(x,0)$ | $q(x,0) \sim \mathcal{N}(0, 25^2(-\Delta_{\text{per}} + 5^2 I)^{-4}))$ |
| Advection | $q_t + a(x)q_x = 0$<br>$q(x,0)=\sin(\pi x), q(0,t)=\sin(\pi t/2)$ | $[0,1]\times[0,1]$ | Transport coefficient $a(x)$ | $g \sim \mathcal{GP}(0,K)$<br>$a(x) = g(x) - \min_{\xi\in[0,1]} g(\xi) + 1$ |

| Darcy | $-\nabla\cdot(a(\boldsymbol{x})\nabla q)=1$ <br> $q\|_{\partial\Omega}=0$ | $(x,y)\in[0,1]^2$ | Permeability field $a(x)$ | $a(\boldsymbol{x})=\exp[g(\boldsymbol{x})]$ <br> $g\sim\mathcal{GRF}_{\cos}(\alpha=2,\tau=3)$ |
|---|---|---|---|---|
| Navier-Stokes | $\omega_t+\boldsymbol{u}\cdot\nabla\omega-\Delta\omega/Re=-4\cos(4y)$ <br> $\nabla\cdot\boldsymbol{u}=0,\omega=v_x-u_y,Re=500$ | $[0,2\pi]^2\times[0,0.5]$ | Initial vorticity field $\omega(\boldsymbol{x},0)$ | $\omega(\boldsymbol{x},0)\sim\mathcal{GRF}_{\text{per}}(\alpha=2.5,\tau=3)$ |

The stochastic condition families in Table S1 are defined as follows. For the one-dimensional Gaussian processes, $K$ denotes the squared-exponential covariance kernel

$$K(x,x')=\exp[-\frac{(x-x')^2}{2\ell^2}],\ell=0.2 \tag{S1}$$

In the Burgers benchmark, $\Delta_{\text{per}}$ denotes the one-dimensional periodic Laplacian. For the Darcy benchmark, $\mathcal{GRF}_{\cos}(\alpha,\tau)$ denotes the truncated cosine Gaussian random field,

$$g(x,y)=\sum_{m=0}^{31}\sum_{n=0}^{31}\xi_{mn}A_{mn}\cos(m\pi x)\cos(n\pi y),\ \xi_{mn}\overset{\text{i.i.d.}}{\sim}\mathcal{N}(0,1) \tag{S2}$$

with spectral amplitudes

$$A_{mn}=\tau^{\alpha-1}\left[\pi^2(m^2+n^2)+\tau^2\right]^{-\alpha/2}c_m c_n,c_0=1,c_{j>0}=\sqrt{2}, \tag{S3}$$

where the constant mode is removed by setting $A_{00}=0$. The removal of the constant mode produces a zero-mean Gaussian field, while the exponential mapping in Table S1 guarantees a strictly positive permeability. For the Navier–Stokes benchmark, $\mathcal{GRF}_{per}(\alpha,\tau)$ denotes a zero-mean periodic Gaussian random field obtained by spectrally filtering a real-valued standard Gaussian white-noise field:

$$\hat{\omega}_{0,\boldsymbol{k}}=\tau^{\alpha-1}\left(|\boldsymbol{k}|^2+\tau^2\right)^{-\alpha/2}\hat{\xi}_{\boldsymbol{k}},\boldsymbol{k}\neq\boldsymbol{0},\hat{\omega}_{0,\boldsymbol{0}}=0 \tag{S4}$$

where $\hat{\xi}_{\boldsymbol{k}}$ denotes the Fourier coefficient of the white-noise field. The compatible initial velocity is recovered from the periodic stream-function according to

$$-\Delta\psi_0=\omega_0,u_0=\partial_y\psi_0,v_0=-\partial_x\psi_0 \tag{S5}$$

thereby ensuring consistency between the initial velocity and vorticity fields and satisfying $\nabla\cdot\boldsymbol{u}_0=0$. In both spectral random-field constructions, $\alpha$ controls the decay of the high-frequency modes, whereas $\tau$ sets the characteristic spectral scale. These condition-generation procedures define the benchmarks used throughout this study and are also used to retrain the corresponding PINO baselines; they are not identical to the Darcy and Navier–Stokes data-generation protocols of the original PINO study.

**S1.2 Physics-informed residual formulation**

Source-model training and subsequent Linearized Subspace Transfer (LST) are both driven by benchmark-specific physics residuals, but they optimize different variables. Source training minimizes the residual loss with respect to the neural-network parameters, whereas LST fixes the trained source

model and its response basis and minimizes a target-condition residual with respect to the response coordinates. For the first five benchmarks, source training and LST use the same continuous PDE-system residual formulation. The Navier–Stokes benchmark instead uses a temporally discretized residual during LST, as detailed in Section S2.4.

Consider a differential-equation system defined on a spatial domain $\Omega$ and, for unsteady problems, a temporal interval $[0,T]$. The unknown scalar- or vector-valued solution is represented by a neural network $q(\boldsymbol{x},t;\boldsymbol{\theta})$, where $\boldsymbol{\theta}$ denotes the trainable parameters. The governing equation and the associated boundary and initial conditions are written as

$$\mathcal{N}[q;\mu]=0, \mathcal{B}[q;\mu]=0, \mathcal{I}[q;\mu]=0 \tag{S6}$$

where $\mu$ denotes the benchmark condition, such as a forcing function, initial condition, or spatially varying coefficient. For steady systems, the temporal variable and the initial-condition operator are omitted.

After evaluating the corresponding operators at their respective sampling points, the governing-equation, boundary-condition, and initial-condition residuals are collected into $\boldsymbol{r}_{\text{PDE}}(q;\mu)\in\mathbb{R}^{N_{\text{PDE}}}, \boldsymbol{r}_{\text{BC}}(q;\mu)\in\mathbb{R}^{N_{\text{BC}}}, \boldsymbol{r}_{\text{IC}}(q;\mu)\in\mathbb{R}^{N_{\text{IC}}}$, where $N_{\text{PDE}}$, $N_{\text{BC}}$ and $N_{\text{IC}}$ denote the corresponding numbers of scalar residual components. A standard PINN minimizes the weighted loss $\mathcal{L}_{\text{PINN}}=\lambda_{\text{PDE}}\mathcal{L}_{\text{PDE}}+\lambda_{\text{BC}}\mathcal{L}_{\text{BC}}+\lambda_{\text{IC}}\mathcal{L}_{\text{IC}}$, with

$$\mathcal{L}_{\text{PDE}}=\frac{1}{N_{\text{PDE}}}\left\|\boldsymbol{r}_{\text{PDE}}(q;\mu)\right\|_2^2, \mathcal{L}_{\text{BC}}=\frac{1}{N_{\text{BC}}}\left\|\boldsymbol{r}_{\text{BC}}(q;\mu)\right\|_2^2, \mathcal{L}_{\text{IC}}=\frac{1}{N_{\text{IC}}}\left\|\boldsymbol{r}_{\text{IC}}(q;\mu)\right\|_2^2 \tag{S7}$$

The relative weights $\lambda_{\text{PDE}}$, $\lambda_{\text{BC}}$ and $\lambda_{\text{IC}}$ balance the contributions of the different residual components. Residual terms that are not applicable to a particular benchmark are omitted.

For consistency with the LST formulation, these weighted residual components are concatenated into a single PDE-system residual vector,

$$\boldsymbol{f}(q;\mu)=\begin{bmatrix}\sqrt{\dfrac{\lambda_{\text{PDE}}N_0}{N_{\text{PDE}}}}\boldsymbol{r}_{\text{PDE}}(q;\mu)\\ \sqrt{\dfrac{\lambda_{\text{BC}}N_0}{N_{\text{BC}}}}\boldsymbol{r}_{\text{BC}}(q;\mu)\\ \sqrt{\dfrac{\lambda_{\text{IC}}N_0}{N_{\text{IC}}}}\boldsymbol{r}_{\text{IC}}(q;\mu)\end{bmatrix}\in\mathbb{R}^{N_0} \tag{S8}$$

where $N_0=N_{\text{PDE}}+N_{\text{BC}}+N_{\text{IC}}$. The PINN loss can therefore be written equivalently as

$$\mathcal{L}_{\text{PINN}}=\frac{1}{N_0}\left\|\boldsymbol{f}(q;\mu)\right\|_2^2 \tag{S9}$$

For a source condition $\mu^{(1)}$, the trained parameter state used as the subsequent linearization point is obtained from

$$\boldsymbol{\theta}_0 = \arg\min_{\boldsymbol{\theta}} \frac{1}{N_0} \left\| \boldsymbol{f}(q(\boldsymbol{x},t;\boldsymbol{\theta});\mu^{(1)}) \right\|_2^2 \tag{S10}$$

The corresponding trained source solution is denoted by $q_0(\boldsymbol{x},t) = q(\boldsymbol{x},t;\boldsymbol{\theta}_0)$.

**S1.3 Source response-space construction and LST implementation**

For each trained source model, the response space is constructed from the output Jacobian evaluated at a fixed set of benchmark-specific basis points. Let

$$\boldsymbol{q}_0 = \begin{bmatrix} q(\boldsymbol{x}_1,t_1;\boldsymbol{\theta}_0) \\ \vdots \\ q(\boldsymbol{x}_{N_b},t_{N_b};\boldsymbol{\theta}_0) \end{bmatrix}, \mathbf{J} = \left.\frac{\partial \boldsymbol{q}}{\partial \boldsymbol{\theta}}\right|_{\boldsymbol{\theta}=\boldsymbol{\theta}_0} \tag{S11}$$

where the temporal coordinate is omitted for steady problems. For vector-valued systems, all output components at the sampled points are stacked into $\boldsymbol{q}_0$. In particular, the Navier–Stokes response space is constructed from the jointly sampled $u$, $v$, and $\omega$ fields. The basis points are fixed for each benchmark and remain unchanged across target conditions.

The dominant right-singular subspace of $\mathbf{J}$ is approximated using the matrix-free randomized SVD procedure introduced in our previous LSR work[3]. The required Jacobian actions are evaluated through Jacobian–vector and vector–Jacobian products, without explicitly forming or storing $\mathbf{J}$. An oversampling dimension of $p = 20$ is used for all benchmarks, and no power iteration is applied. The parameter directions are ordered according to decreasing approximate singular value, and the first $r$ directions are retained: $\mathbf{V} = [\boldsymbol{v}_1, \boldsymbol{v}_2, \ldots, \boldsymbol{v}_r]$. The corresponding response modes are evaluated as $\boldsymbol{\Phi}(\boldsymbol{x},t) = \mathbf{J}(\boldsymbol{x},t)\mathbf{V}$.

After constructing $\mathbf{V}$, the source prediction, response modes and all coordinate derivatives required by the corresponding PDE system are evaluated and stored as a PDE-ready response cache. The cached quantities include both the source field and the corresponding response modes. The target residual and its response-coordinate Jacobian can therefore be assembled directly from algebraic combinations of the cached quantities. Online LST consequently requires neither reevaluation nor coordinate or parameter differentiation of the source network on the prescribed residual and output grids.

For the target LST problems, the response coordinates are initialized as $\boldsymbol{\alpha}^{(0)} = \boldsymbol{0}$.

At the $i$-th Gauss-Newton iteration, the response-coordinate residual and its Jacobian are evaluated as $\boldsymbol{f}^{(i)} = \boldsymbol{f}\left(\boldsymbol{q}_0 + \boldsymbol{\Phi}\boldsymbol{\alpha}^{(i)};\mu\right)$, and

$$\mathbf{G}^{(i)} = \left.\frac{\partial \boldsymbol{f}\left(\boldsymbol{q}_0 + \boldsymbol{\Phi}\boldsymbol{\alpha}; \mu\right)}{\partial \boldsymbol{\alpha}}\right|_{\boldsymbol{\alpha}=\boldsymbol{\alpha}^{(i)}} \tag{S12}$$

The response-coordinate residual Jacobian is assembled analytically from the cached source fields and modal derivatives according to the governing equations, without reevaluating or differentiating the source network during online transfer.

The update is obtained by solving

$$\Delta\boldsymbol{\alpha}^{(i)} = \arg\min_{\Delta\boldsymbol{\alpha}} \left\| \mathbf{G}^{(i)}\Delta\boldsymbol{\alpha} + \boldsymbol{f}^{(i)} \right\|_2^2 \tag{S13}$$

followed by $\boldsymbol{\alpha}^{(i+1)} = \boldsymbol{\alpha}^{(i)} + \Delta\boldsymbol{\alpha}^{(i)}$.

For PDE systems that are linear in the solution field, the residual is affine in $\boldsymbol{\alpha}$, and a single direct least-squares solve is sufficient. For nonlinear systems, multiple Gauss-Newton updates are applied. The benchmark-specific ranks, iteration counts, numerical precisions, and least-squares settings are provided in Section S3.

## S2. Methodological choices and sensitivity analyses

### S2.1 Effect of source-training configuration on transfer

To examine how source-model optimization affects subsequent transfer, we trained four Burgers source models using Adam32, L-BFGS32, SSBroyden32, and SSBroyden64. Adam32 and L-BFGS32 represent first-order and limited-memory quasi-Newton baselines, respectively, whereas SSBroyden32 and SSBroyden64 denote the full-memory self-scaled Broyden optimizer operated in single and double precision. Following Ref.[4], both SSBroyden runs were initialized from the model obtained after 1,000 Adam iterations. Because full-memory Broyden optimization maintains a dense approximation of second-order information, all four models in this comparison used the same compact fully connected network with four hidden layers of width 20 and tanh activations.

Figure S1 compares the convergence histories of the physics-informed loss and solution error. Adam32 reduces both quantities only gradually and remains at a comparatively high error level throughout training. L-BFGS32 achieves a more sustained decrease in the loss and error, yielding a substantially more accurate source model with the same network architecture. Starting from the model obtained after 1,000 Adam iterations, both SSBroyden runs produce a further sharp reduction in error and attain markedly higher accuracy than the Adam32 and L-BFGS32 baselines. SSBroyden32 eventually reaches a single-precision plateau, whereas SSBroyden64 continues to converge to a substantially lower error level. The resulting source-model errors span more than three orders of magnitude, providing a controlled basis for assessing how source optimization influences subsequent

LST.

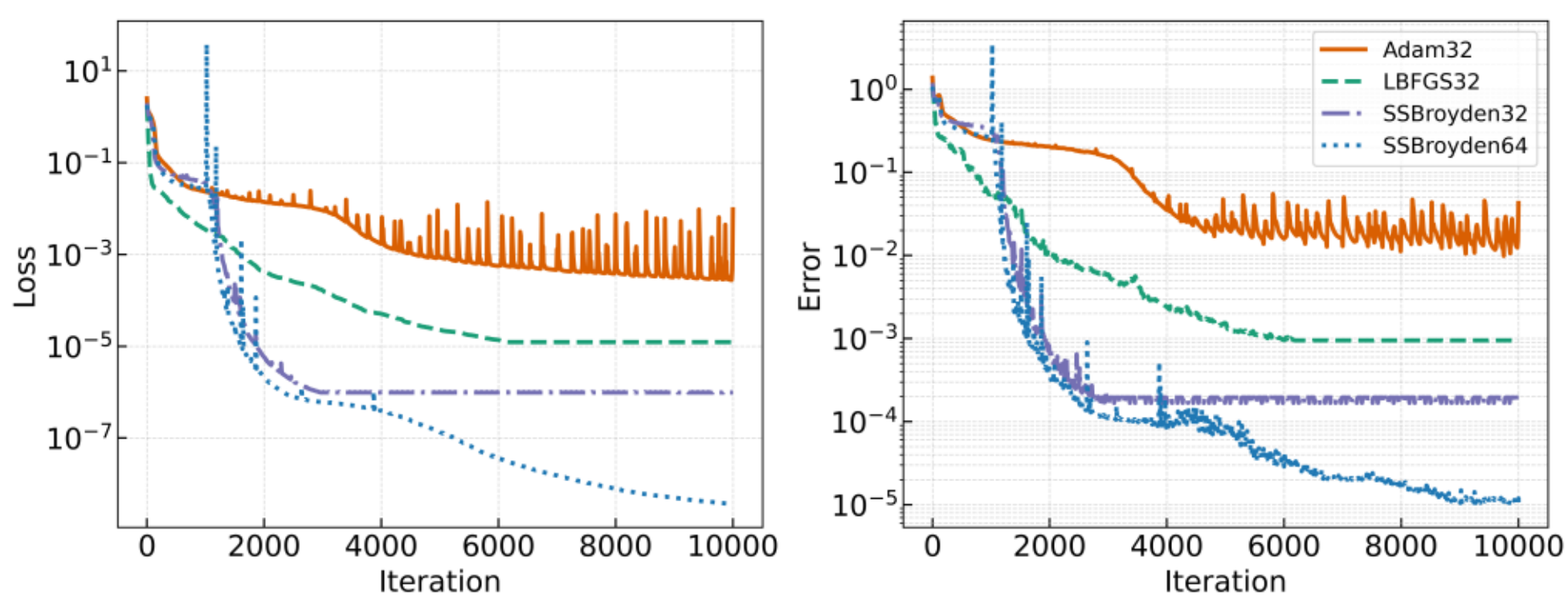

Fig. S1. Convergence of Burgers source models under different optimizers and numerical precisions.

Each trained model was then used as the source for transfer to the same representative target condition. The corresponding source and target reference solutions are shown in Fig. 2c, e of the main text, and Fig. S2 reports the transfer loss and error as functions of response-space rank. For source models originally trained in single precision, double-precision LST was performed by first casting the trained parameters to double precision and then constructing the response space and PDE-ready cache in double precision.

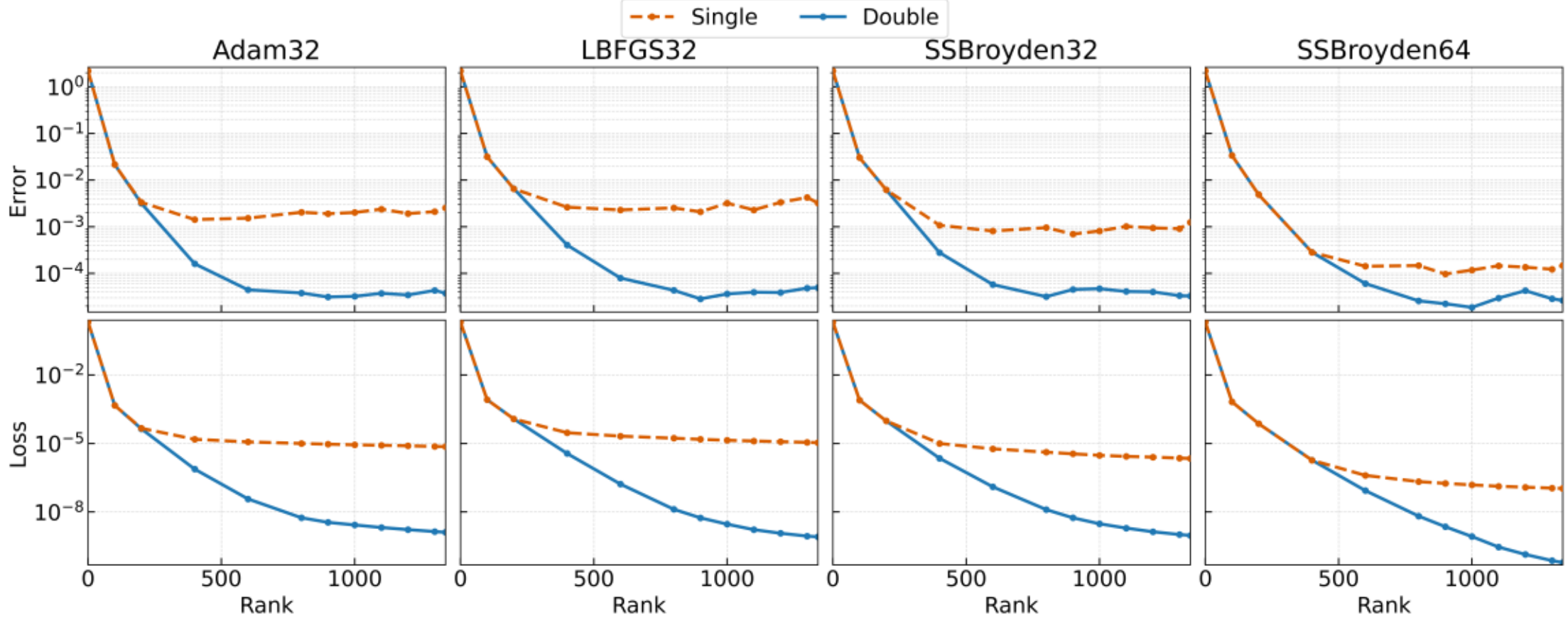

Fig. S2. Rank-dependent Burgers transfer performance for source models obtained using different optimizers and numerical precisions.

At $r = 0$, LST reduces to the uncorrected source prediction, yielding target errors close to unity for all four models. Increasing the rank rapidly reduces both the transfer loss and solution error, demonstrating that the source-derived response spaces contain directions capable of representing the source-to-target discrepancy. At lower ranks, the single- and double-precision results are nearly indistinguishable, indicating that the error is primarily limited by the approximation capacity of the truncated response space. At larger ranks, the two precision settings separate as the response-coordinate problem becomes increasingly sensitive to ill-conditioning and finite-precision effects.

Double precision is therefore required in this setting to exploit the higher-rank response directions and attain the lowest transfer errors.

Despite the large differences in source-model accuracy, the four response spaces yield comparable transfer-error levels at sufficiently high rank when LST is performed in double precision. This indicates that source-condition error alone does not determine the attainable target accuracy. Once optimization has produced a sufficiently informative local response structure, transfer performance is governed primarily by how well that response space represents the target discrepancy and by the numerical stability of the response-coordinate solve. Taken together, these results identify L-BFGS32 as a practical default for source-model training. It produces substantially more accurate source models than Adam32 while retaining markedly better scalability than the full-memory SSBroyden optimizers. Accordingly, unless otherwise stated, L-BFGS32 is used to train the source models in all subsequent experiments.

Table S2 extends this comparison to 100 target conditions using $r = 1,000$ and double-precision LST. Although the source errors span more than three orders of magnitude, the median transfer errors remain tightly clustered between approximately $2\times10^{-5}$ and $3\times10^{-5}$. This result should not be interpreted as implying that source optimization is unimportant. As shown in Fig. 2h of the main text, source training substantially strengthens the transferable response representation for Burgers. Rather, once this structure has formed, further reductions in source-condition error do not necessarily translate into proportional improvements in typical target-condition accuracy.

For each source model, the mean transfer error exceeds the median, indicating a right-skewed distribution in which a small number of difficult targets disproportionately affect the average. The mean and standard deviation therefore characterize sensitivity to tail cases, whereas the median and interquartile range describe transfer performance for a typical target. The mean online adaptation time is approximately 0.46 s for all four source models. Once the response information has been cached, the online cost is governed mainly by the response-coordinate solve at the prescribed rank and is therefore largely independent of the optimizer used during offline source training.

Table S2. Burgers transfer performance for source models trained using different optimizers. Results are evaluated over 100 target conditions using a response-space rank of $r=1,000$ and double-precision LST. The interquartile range is reported as [Q1, Q3].

| | Adam32 | L-BFGS32 | SSBroyden32 | SSBroyden64 |
|---|---|---|---|---|
| Source Error | 2.56e-2 | 9.49e-4 | 1.95e-4 | 1.06e-5 |
| Transfer Error (mean ± s.d.) | 2.09e-4 ± 1.10e-3 | 2.14e-4 ± 1.50e-3 | 3.00e-4 ± 1.90e-3 | 1.12e-3 ± 7.3e-3 |
| Transfer Error (median [Q1, Q3]) | 2.61e-5 [1.89e-5,6.00e-5] | 1.96e-5 [1.33e-5,4.62e-5] | 2.65e-5 [1.40e-5,6.73e-5] | 2.93e-5 [1.82e-5,1.14e-4] |

We further examined whether the response structure underlying this transfer behavior is specific to the optimizer or network architecture. We compared rank-matched response spaces obtained from the four optimizer configurations above and from Adam-trained networks of different widths (Fig. S3). Response-space similarity was quantified using the mean of the squared cosines of their principal angles. The spaces exhibit substantial overlap across both optimizer and width variations, particularly among their leading directions, although their detailed higher-rank components are not identical.

Despite these geometric differences, the corresponding affine projection errors follow similar rank-dependent trends across the target family. Both the mean and upper-tail projection errors decrease as additional response modes are retained, indicating comparable capacity to represent cross-condition solution variation. Together with the transfer results in Table S2, these observations show that the useful response structure is not specific to a particular optimizer or network width, and that source-model accuracy alone does not characterize the transfer capacity of the resulting response space.

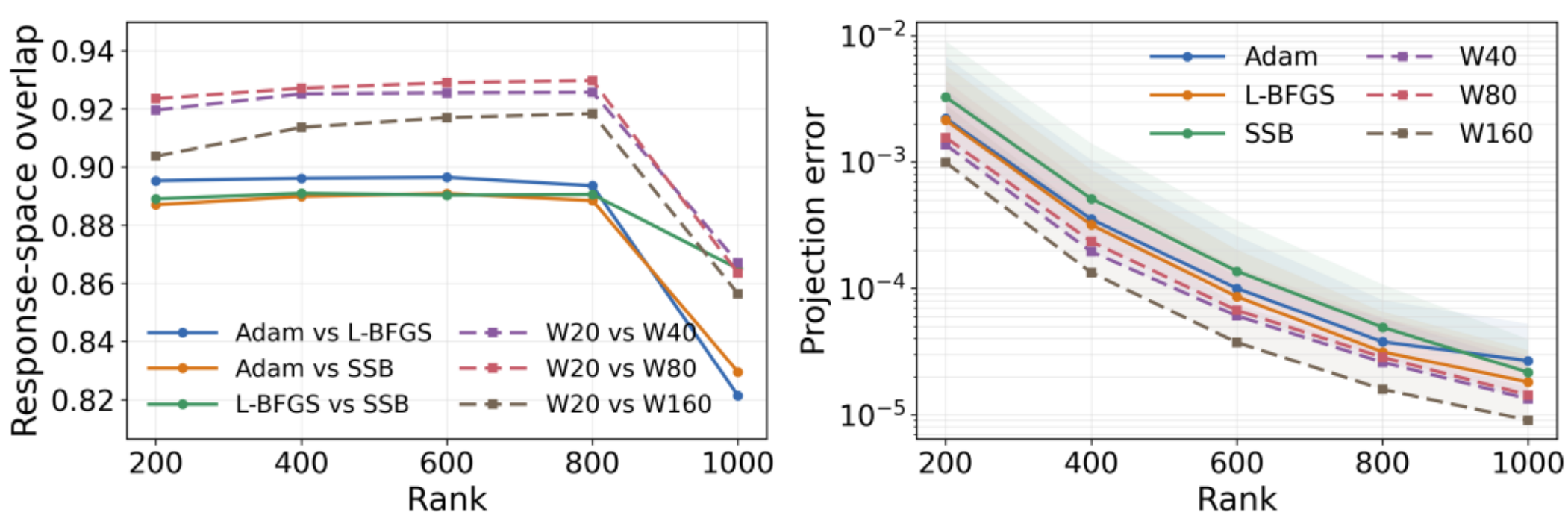

Fig. S3. Robustness of Burgers response spaces across source-training configurations. Left: Pairwise overlap between rank-matched response spaces obtained using different optimizers and network widths, defined as the mean of the squared cosines of their principal angles. Width comparisons use the width-20 Adam model as the reference. Right: Affine projection errors of the Burgers target family onto the corresponding response spaces. Lines show the mean and shading extends from the mean to the 90th percentile. All projection errors are relative $L_2$ errors.

### S2.2 Response-space optimization and full-parameter target adaptation

The rapid target adaptation of LST may arise either from the trained source parameters providing a favorable initialization or from the response-space formulation itself. To distinguish these effects, we first performed full-parameter target optimization for the representative Burgers condition used in Fig. 2a–c of the main text. The target physics-informed objective was minimized using Adam32, L-BFGS32, SSBroyden32 and SSBroyden64, starting either from random initialization or from the parameters of the trained source model. The network architecture, target residual formulation, residual samples, numerical precision and optimizer settings were otherwise identical between the two initialization strategies.

Source initialization modifies the early optimization transient but does not fundamentally change the optimizer-dependent convergence behavior (Fig. S4). The warm-started and from-scratch runs approach similar error scales under the same optimizer, and source initialization does not consistently accelerate convergence over the full optimization trajectory. In particular, achieving errors substantially below the single-precision plateaus still requires prolonged target-specific full-parameter optimization. Thus, the rapid adaptation observed with LST cannot be explained simply by warm-starting the target optimization from a trained source network.

We next optimized the rank-1,000 affine response-space model using the same four generic optimizers rather than the nonlinear least-squares solver used for the main LST results (Fig. S5). The resulting convergence remains strongly optimizer dependent and is qualitatively similar to full-parameter optimization. This comparison highlights that the response-space restriction alone does not make generic first-order or quasi-Newton optimization converge in only a few iterations. The computational advantage of LST instead arises from the combination of a fixed affine solution representation and an optimization procedure that explicitly exploits the resulting least-squares structure.

Once the response space is constructed, the target solution takes the form $q = q_0 + \boldsymbol{\Phi}\boldsymbol{\alpha}$, so the source network no longer needs to be differentiated during target adaptation. The response modes and their required spatial and temporal derivatives can therefore be precomputed and cached. For nonlinear PDEs, target adaptation becomes a structured nonlinear least-squares problem over the response coordinates, for which each Gauss–Newton or Levenberg–Marquardt step reduces to a linear least-squares solve; for linear differential operators, the target residual is affine in the response coordinates and can be solved directly by linear least squares. The structure-exploiting least-squares formulation enables the few-step convergence observed in Fig. 2a, whereas caching the response fields and their PDE-relevant derivatives substantially reduces the cost of each residual evaluation.

As an implementation-level control, optimizing the rank-1,000 response model with Adam required 11.44 s when the response modes and their derivatives were cached, whereas a differentiation-based implementation that did not exploit this fixed representation required 619.19 s under the corresponding optimization protocol. Together with the full-parameter controls above, these results show that the efficiency of LST derives from solving the target physics through structure-exploiting least-squares updates in a fixed affine response representation whose fields and PDE-relevant derivatives can be reused throughout adaptation.

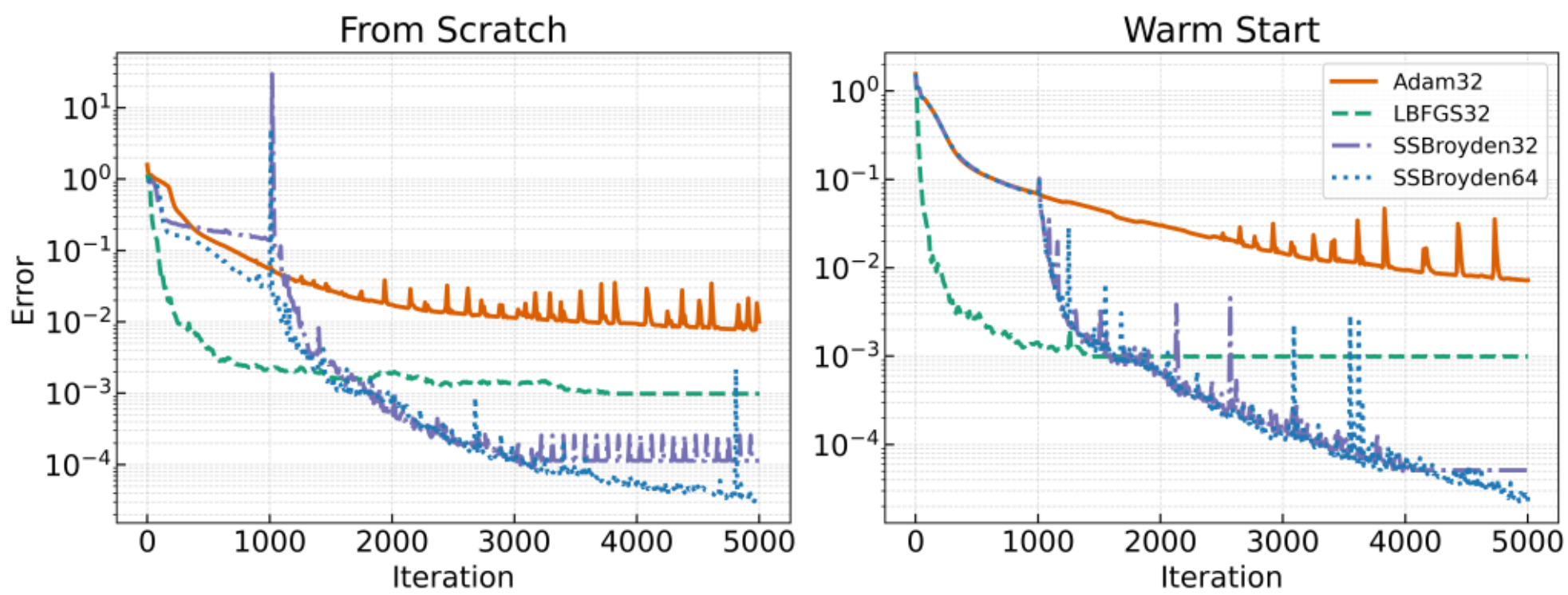

Fig. S4. Full-parameter optimization at a representative Burgers target condition from random and source-model initialization.

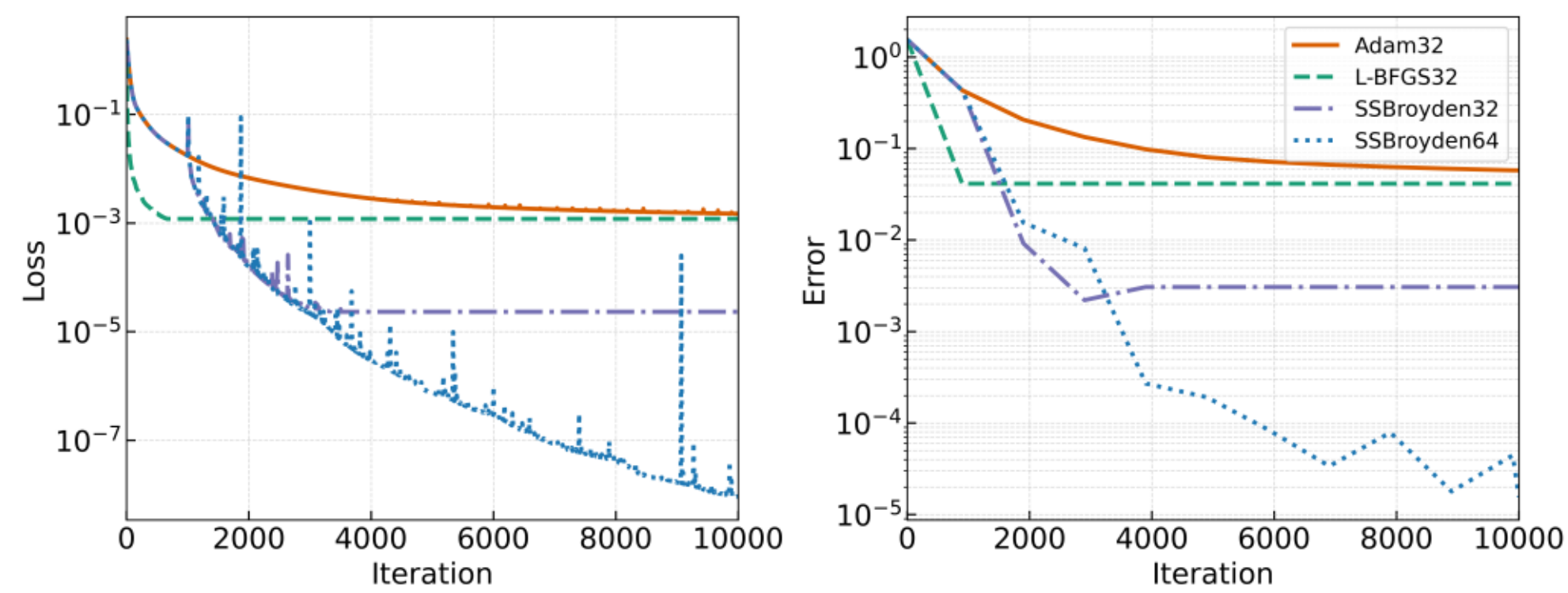

Fig. S5. Optimization of a fixed rank-1,000 response-space model using generic optimizers.

### S2.3 Robustness to response-space rank and fixed-rank selection

The response-space rank controls the trade-off between the approximation capacity of the source-derived response space and the numerical conditioning of the response-coordinate solve. In this work, the rank is treated as a global method-level hyperparameter: a single value is fixed for all source models, candidate conditions, ATM enrichment stages, and unseen test conditions within a benchmark. The rank is not adjusted for individual target conditions. To assess the sensitivity of Burgers transfer to this hyperparameter, we retrospectively evaluated $r \in \{400, 600, 800, 1000, 1200, 1341\}$, where $r = 1{,}341$ corresponds to the full number of trainable parameters in the source network. The analysis was performed over the 100 Burgers candidate conditions. Reference solutions at these conditions are used only in this section to diagnose rank sensitivity; they are not used in the LST objective or in ATM source acquisition, routing, stopping, or target-time deployment.

For diagnostic purposes, we define the oracle condition-specific optimal rank as the tested rank yielding the lowest reference-solution error for a given condition. As summarized in Table S3, increasing the rank from 400 to 1,000 consistently improves both typical and upper-tail transfer accuracy. The median error decreases from $9.01\times10^{-5}$ at $r = 400$ to $1.30\times10^{-5}$ at $r = 1{,}000$, while the 90th-percentile error decreases by nearly two orders of magnitude, from $1.38\times10^{-2}$ to $1.82\times10^{-4}$.

The improvement is particularly pronounced for the more difficult target conditions, indicating that insufficient rank primarily limits the coverage of the source response space.

The oracle candidate-specific optimal ranks are strongly concentrated at $r = 800$ and $r = 1,000$. Among the 100 candidates, 19 attain their lowest error at $r = 800$, whereas 71 attain it at $r = 1,000$; these two ranks therefore account for 90% of all candidate-wise optima. Their median errors, $1.52\times10^{-5}$ and $1.30\times10^{-5}$, respectively, are also close, indicating that the typical transfer accuracy is relatively insensitive within this favorable rank range. Nevertheless, $r = 1,000$ provides substantially better upper-tail performance, reducing the 90th-percentile error from $6.70\times10^{-4}$ at $r = 800$ to $1.82\times10^{-4}$. Moreover, for 86% of the candidates, the error obtained at the fixed rank $r = 1,000$ is within 10% of the minimum error attained over all tested ranks. The broad plateau indicates that rank selection does not require target-specific tuning or precise identification of a sharp optimum.

Table S3. Post hoc sensitivity of Burgers transfer performance to response-space rank. Statistics are evaluated over 100 candidate conditions. The final row reports the number of candidates for which each tested rank produces the lowest transfer error.

| Rank | 400 | 600 | 800 | 1,000 | 1,200 | 1,341 |
|---|---|---|---|---|---|---|
| Median error | 9.01e-5 | 2.68e-5 | 1.52e-5 | 1.30e-5 | 2.33e-5 | 2.83e-5 |
| P90 error | 1.38e-2 | 2.77e-3 | 6.70e-4 | 1.81e-4 | 1.91e-4 | 2.11e-4 |
| Oracle-optimal count | 0 | 0 | 19 | 71 | 5 | 5 |

**S2.4 Selection of the Navier–Stokes temporal representation**

The Navier–Stokes benchmark raises an additional question concerning the temporal structure of the response-coordinate representation. In the global space–time formulation used for the preceding benchmarks, a single response-coordinate vector is optimized over the entire space–time domain. For complex unsteady flow dynamics, however, requiring the same coordinates to represent all temporal states may impose a severe restriction. We therefore performed an oracle projection analysis to distinguish this representational limitation from errors introduced by the physics-constrained online solve.

In the global formulation, the reference trajectory was fitted using a single response-coordinate vector over either 20,000 or 30,000 randomly sampled space–time points. In the time-local formulation, an independent coordinate vector was fitted at each time level. Two time-local response spaces were considered: a dynamic representation using the source response modes evaluated at the current time, $\mathbf{\Phi}(t)$, and a frozen representation that reuses the initial-time modes, $\mathbf{\Phi}_0 = \mathbf{\Phi}(0)$, throughout the trajectory. Because these fits directly access the reference solution, they quantify the approximation capacity of each representation independently of the PDE residual and time-marching algorithm.

As summarized in Table S4, the global projection error decreases as the rank increases, but remains substantial even at $r = 5,000$. Increasing the number of fitting points from 20,000 to 30,000 changes the mean error only from 6.365% to 6.052%, indicating that the global projection has already become largely insensitive to the fitting-point count. Its remaining error therefore reflects a representational limitation rather than insufficient sampling. By contrast, allowing the response coordinates to vary with time reduces the mean oracle error to 0.358% at the same rank. The corresponding P90 errors are 6.570% for the global formulation and 0.376% for the time-local dynamic representation. These results identify the use of a single global coordinate vector over the complete space–time domain as the dominant representation bottleneck.

Once a sufficiently large rank is used, dynamically updating the response modes provides little additional benefit over reusing the initial-time modes. At $r = 3,000$ , the dynamic and frozen formulations yield mean errors of 1.286% and 1.292%, respectively. At $r = 5,000$ , the difference further decreases to only 0.005 percentage points, with mean errors of 0.358% and 0.363%. Their P90 errors are similarly close, at 0.376% and 0.387%. Thus, although the response coordinates must evolve in time, the underlying response basis need not be reconstructed at every time level once its rank is sufficiently large.

These observations motivate the formulation adopted in the formal Navier–Stokes experiments: the response coordinates are advanced sequentially in time, while the initial-time response modes $\mathbf{\Phi}$ are constructed once, cached, and reused at all subsequent time steps. This design removes the restrictive global space–time coordinate assumption without requiring repeated response-space construction during online transfer.

Table S4. Oracle projection error for global and time-local Navier–Stokes response representations. Values are mean errors, reported as mean ± s.d.

| Rank | Global, 20k points | Global, 30k points | Time-local, dynamic $\mathbf{\Phi}(t)$ | Time-local, frozen $\mathbf{\Phi}_0$ |
|---|---|---|---|---|
| 1000 | 0.17251 ± 0.00436 | 0.17114 ± 0.00431 | 0.07315 ± 0.00151 | 0.08404 ± 0.00244 |
| 3000 | 0.08612 ± 0.00361 | 0.08384 ± 0.00353 | 0.01286 ± 0.00014 | 0.01292 ± 0.00018 |
| 5000 | 0.06365 ± 0.00299 | 0.06052 ± 0.00293 | 0.00358 ± 0.00007 | 0.00363 ± 0.00010 |

These results motivate the temporal formulation used in the formal Navier–Stokes experiments. Specifically, the response coordinates are allowed to evolve in time, whereas the initial-time response basis is constructed once and reused throughout the trajectory. At time level $t_n$ , the transferred state is represented as $\boldsymbol{q}_n = \boldsymbol{q}_{s,n} + \mathbf{\Phi}_0 \boldsymbol{\alpha}_n$ , where $\boldsymbol{q}_{s,n}$ is the source trajectory at $t_n$ . The initial response coordinates are obtained by ridge-regularized projection of the prescribed target initial condition:

$$\boldsymbol{\alpha}_0 = \arg\min_{\boldsymbol{\alpha}} \left\| \boldsymbol{q}_0 + \mathbf{\Phi}_0 \boldsymbol{\alpha} - \boldsymbol{q}_0^{\mathrm{tar}} \right\|_2^2 + \lambda_{\mathrm{IC}} \left\| \boldsymbol{\alpha} \right\|_2^2 \tag{S14}$$

with $\lambda_{\mathrm{IC}} = 10^{-2}$ . The ridge term suppresses poorly conditioned response directions and prevents excessively large initial coordinates that could amplify errors during subsequent time marching.

The response coordinates are subsequently advanced over 64 Crank–Nicolson time steps. At each step, one Levenberg–Marquardt-regularized Gauss–Newton update is applied to the time-discrete residual:

$$(\mathbf{G}_n^T\mathbf{G}_n + \gamma\mathbf{I})\Delta\boldsymbol{\alpha}_n = -\mathbf{G}_n^T\boldsymbol{f}_n \tag{S15}$$

where $\boldsymbol{f}_n$ is the Crank-Nicolson residual and $\mathbf{G}_n$ is its response-coordinate Jacobian. The damping term improves the conditioning of the local Gauss-Newton system and limits updates along weakly constrained response directions. Because only one update is performed per time step, it also controls the effective magnitude of the response-coordinate evolution. We use $\gamma = 10$ and one update per time step for all transferred conditions. Neither regularization term modifies the governing equations or the Crank-Nicolson residual; they act only on the identification and evolution of the response coordinates. No target-specific tuning is performed.

# S3. Numerical settings and baseline configurations

## S3.1 Source-model and transfer settings for ATM benchmarks

All experiments were performed on a single NVIDIA GeForce RTX 5080 GPU. All reported wall-clock costs correspond to single-GPU execution without model or data parallelism. The default source-model and transfer configurations used in the main ATM experiments are summarized in Table S5. The settings in Table S5 are treated as benchmark-level hyperparameters and are held fixed across all sources, candidate conditions, enrichment stages and unseen test conditions within each benchmark; no target-specific or test-set-based tuning is performed. In particular, reference errors on the unseen test set are not used to select or adjust these settings. Unless otherwise stated, all source models were trained in single precision using L-BFGS. The residual blocks were first normalized by their respective numbers of scalar components, as defined in Section S1.2, and were then combined using benchmark-specific relative weights. For the first four benchmarks, these weights follow the corresponding settings of Wang et al.[1], whereas the Darcy and Navier–Stokes source-model and transfer configurations were specified for the present study.

Table S5. Default source-model and LST configurations for the six benchmark systems. The notation w×d denotes a fully connected network with d hidden layers of width w. Residual counts refer to source-model training. A periodic-point pair in the Burgers benchmark contributes separate residuals for the solution and its first spatial derivative.

| System | Source network | Residual samples | Relative weights | Response-space rank | Response-coordinate updates |
|---|---|---|---|---|---|
| Linear ODE | 20×4 tanh MLP | 1 BC + 1,000 PDE | $(\lambda_{\mathrm{BC}}, \lambda_{\mathrm{PDE}}) = (1,1)$ | 40 | 1 |
| Diffusion-reaction | 20×4 tanh MLP | 2,000 IC/BC +15,000 PDE | $(\lambda_{\mathrm{IC/BC}}, \lambda_{\mathrm{PDE}}) = (1,1)$ | 1,000 | 3 |

| Burgers | 20×4 tanh MLP | 101 IC + 1,000 BC + 10,000 PDE | $(\lambda_{\mathrm{IC}}, \lambda_{\mathrm{BC}}, \lambda_{\mathrm{PDE}}) = (20,1,1)$ | 1,000 | 5 |
|---|---|---|---|---|---|
| Advection | 20×4 tanh MLP | 1,000 IC/BC + 10,000 PDE | $(\lambda_{\mathrm{IC/BC}}, \lambda_{\mathrm{PDE}}) = (100,1)$ | 1,000 | 1 |
| Darcy | 32×6 tanh MLP | 20,000 PDE | $\lambda_{\mathrm{PDE}} = 1$ | 2,500 | 1 |
| Navier-Stokes | 32×6 tanh MLP | 3,000 IC + 20,000 PDE | $(\lambda_{\mathrm{IC}}, \lambda_{\mathrm{PDE}}) = (10,1)$ | 5,000 | 1 per time step |

For Darcy flow, the homogeneous Dirichlet boundary condition is imposed exactly through the hard-constrained output representation $q(\boldsymbol{x};\boldsymbol{\theta}) = x(1-x)y(1-y)\,\hat{q}(\boldsymbol{x};\boldsymbol{\theta})$, thereby eliminating the need for boundary residual points. For the Navier–Stokes benchmark, we adapt a previously proposed input representation[5]. The normalized time coordinate and the two spatial coordinates are first represented as $z = [t/T, \cos x, \sin x, \cos y, \sin y]$, which enforces the 2π-periodicity of the predicted fields by construction. The resulting vector is then mapped through a trainable 256-dimensional Fourier feature embedding before being passed to the six-layer, width-32 tanh MLP. The network predicts the two velocity components $(u, v)$, while the vorticity is obtained consistently as $\omega = v_x - u_y$.

For the first five benchmarks, source training and LST use the same benchmark-specific residual definitions and relative weights. The Navier-Stokes benchmark uses a separate time-discrete residual for LST, as detailed in Section S2.4. For each benchmark, one randomly sampled condition is used to train the initial source model, while the candidate and test sets each contain 100 non-overlapping conditions drawn from the same underlying distribution. The candidate set is used exclusively for active source acquisition and stopping, whereas the independent test set is reserved for evaluating generalization to unseen conditions. Reference solutions are used only for post hoc error assessment and do not enter source acquisition, stopping, or routing.

### S3.2 Physics-informed operator-learning baseline

For the operator-learning comparisons, we use the publicly released implementations associated with Wang et al.[1] for the linear ODE, diffusion–reaction, Burgers and advection benchmarks, and the PINO implementation of Li et al.[2], retrained on our benchmark definitions, for Darcy flow and Navier–Stokes. All reported baseline errors and wall-clock times are obtained from our own runs on the same computing platform used for ATM, rather than taken from the original publications.

For the first four benchmarks, we follow the source implementations of Wang et al.[1], reimplementing the models in PyTorch while preserving the original network architectures, condition distributions, physics-informed objectives, collocation strategies and optimization settings. The resulting models therefore provide closely matched physics-informed operator baselines for these systems. The source implementation[1] optimizes PI-DeepONet using Adam, and the resulting model is therefore used as the primary Burgers operator baseline. Training is continued for 200,000 optimization steps and requires 6.6 h on our hardware. Both the physics-informed loss and mean test error largely plateau after approximately 100,000 steps, corresponding to about 3 h of training, with the mean relative test error remaining near $3.96\times10^{-2}$. To assess whether the comparison is limited by the optimizer, we additionally train the same model using L-BFGS while keeping the architecture, training conditions and physics-informed objective unchanged. The L-BFGS model reaches a lower mean relative test error of $1.12\times10^{-2}$, but requires approximately 8 h (Fig. S6). Initial-condition, boundary and residual points are resampled at each outer epoch, and the optimizer history is reset for each new physics batch. The reported checkpoint is obtained after 350 outer epochs, corresponding to 177,182 closure evaluations. This additional result is used only as an optimizer sensitivity test and is not used as the primary value in Table 1.

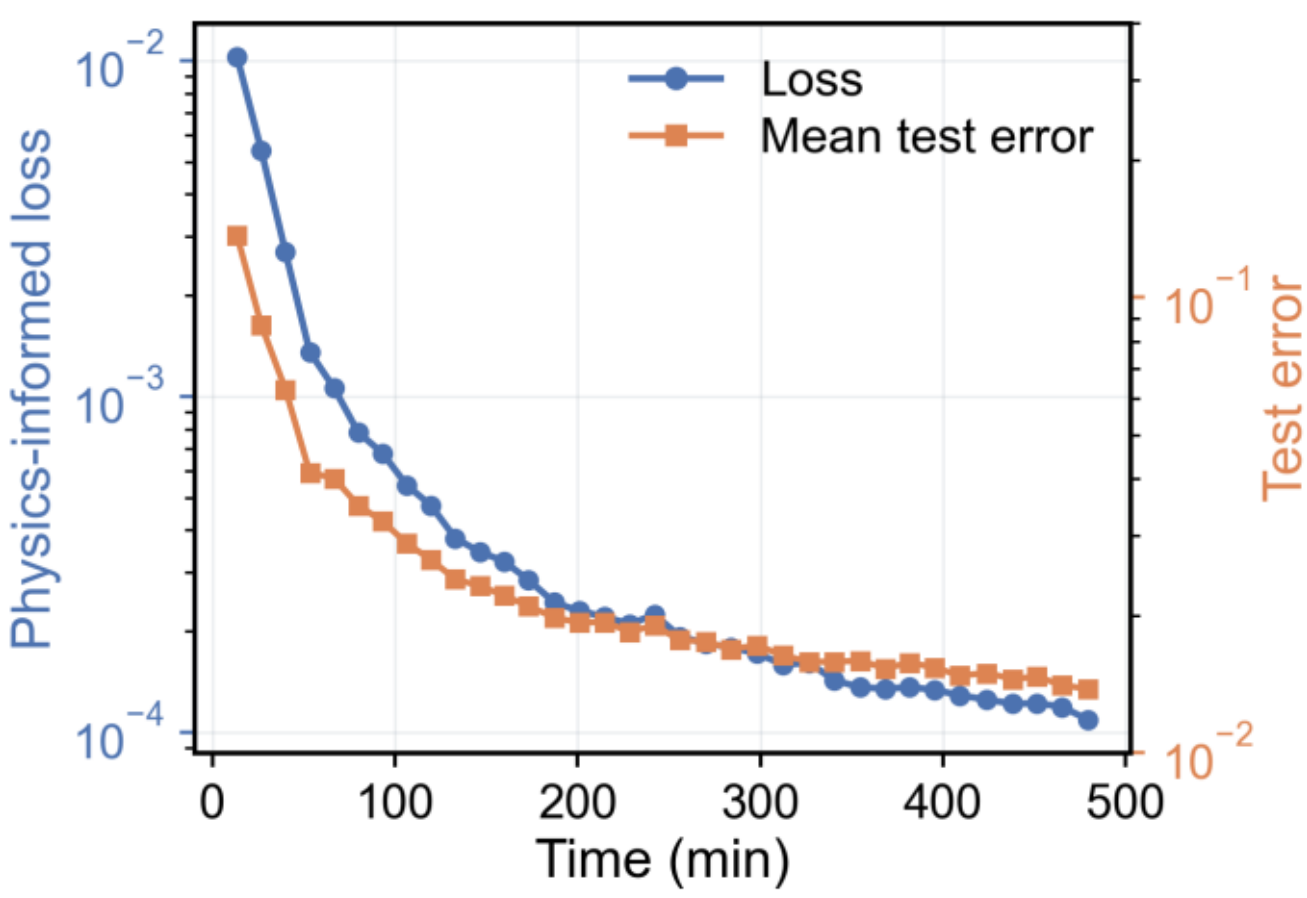

Fig. S6. L-BFGS convergence history for the Burgers PI-DeepONet.

For Darcy flow and Navier–Stokes, we use the publicly released PINO implementation of Li et al.[2], retrained and evaluated under exactly the same benchmark definitions used for ATM. These benchmark definitions differ from those of the original PINO study. For Darcy flow, we replace the original permeability-field distribution with the coefficient-field family defined in Section S1.1. For Navier–Stokes, initial vorticity fields are sampled directly from the prescribed periodic Gaussian random field and used as the starting states for evolution; unlike the original PINO data-generation protocol, no long forced burn-in is applied to generate attractor-state initial conditions. The PINO architecture and its associated grid-based differentiation, residual discretization and optimization pipeline otherwise follow the released implementation. Because these components differ from the PI-DeepONet baselines used for the other four systems, the Darcy and Navier–Stokes results are treated as reference physics-informed operator comparisons rather than directly matched counterparts to those four baselines. All reported errors and wall-clock costs are obtained from our own runs on the same computing platform.

## S4. Robustness of single-source transfer and final ATM performance

### S4.1 Robustness of LST to source selection

To determine whether the cross-condition transfer capability of LST depends on a particularly favorable source condition, we repeated the single-source experiments using three independently sampled source conditions for each benchmark. These additional sources are distinct from both the initial source used in the formal ATM experiments and the 100 unseen target conditions used for evaluation. Each source model was trained independently using the same benchmark-specific protocol, after which its response space was constructed and evaluated on the same target set. The network architecture, response-space rank, target set, residual sampling, numerical precision and response-coordinate solver were held fixed, with no source-specific tuning. For each source, we additionally constructed a matched initialization-state response representation from the random parameter state preceding source training.

Table S6. Robustness of single-source LST across independently sampled source conditions and matched initialization states. Three additional sources are evaluated for each benchmark over the same 100 unseen targets. Source error denotes the trained-model error at the source condition. Init. LST uses the response representation constructed at the matched random initialization. Gain is the ratio of the Init.-LST mean error to the trained-LST mean error; values greater than one favour training. R1–R3 exclude the initial source used in the formal ATM experiments.

| | Source | Source-model error | Trained LST | | | | Init. LST | Gain |
|---|---|---|---|---|---|---|---|---|
| | | | Mean | Median | P90 | Max | Mean | |
| Linear ODE | R1 | 1.07e-4 | 1.09e-6 | 6.07e-7 | 2.15e-6 | 9.50e-6 | 2.73e-6 | 2.5× |
| | R2 | 5.08e-4 | 9.68e-8 | 7.60e-8 | 2.05e-7 | 3.39e-7 | 3.90e-6 | 40.3× |
| | R3 | 1.72e-4 | 9.71e-8 | 7.64e-8 | 2.08e-7 | 3.47e-7 | 2.84e-6 | 29.2× |
| Diffusion-reaction | R1 | 1.57e-3 | 1.03e-4 | 9.41e-5 | 1.78e-4 | 3.82e-4 | 3.19e-4 | 3.1× |
| | R2 | 3.50e-3 | 2.14e-4 | 1.86e-4 | 3.59e-4 | 9.03e-4 | 2.41e-4 | 1.1× |
| | R3 | 6.16e-3 | 1.37e-4 | 1.28e-4 | 2.41e-4 | 3.89e-4 | 2.40e-4 | 1.8× |
| Burgers | R1 | 2.25e-2 | 3.79e-4 | 2.22e-5 | 1.14e-3 | 9.31e-3 | 3.11e-2 | 82.0× |
| | R2 | 1.61e-3 | 2.79e-4 | 3.10e-5 | 1.57e-4 | 1.12e-2 | 8.52e-3 | 30.6× |
| | R3 | 9.46e-4 | 1.87e-4 | 9.30e-6 | 5.65e-5 | 7.47e-3 | 1.45e-2 | 77.2× |
| Advection | R1 | 7.69e-3 | 2.58e-3 | 2.38e-3 | 3.87e-3 | 6.07e-3 | 1.34e-2 | 5.2× |
| | R2 | 9.56e-3 | 3.24e-3 | 3.20e-3 | 4.26e-3 | 6.54e-3 | 1.37e-2 | 4.2× |
| | R3 | 3.69e-3 | 4.80e-3 | 4.65e-3 | 7.28e-3 | 9.31e-3 | 1.13e-2 | 2.3× |
| Darcy | R1 | 8.77e-3 | 3.85e-4 | 3.16e-4 | 7.23e-4 | 1.40e-3 | 4.70e-2 | 122.1× |
| | R2 | 9.51e-3 | 2.70e-4 | 2.26e-4 | 4.89e-4 | 8.67e-4 | 4.63e-2 | 171.4× |
| | R3 | 6.81e-3 | 2.25e-4 | 2.06e-4 | 3.75e-4 | 6.22e-4 | 4.56e-2 | 203.0× |
| Navier-Stokes | R1 | 3.89e-2 | 4.22e-2 | 4.07e-2 | 5.18e-2 | 6.92e-2 | 7.12e-2 | 1.7× |
| | R2 | 3.96e-2 | 4.97e-2 | 4.72e-2 | 6.52e-2 | 8.07e-2 | 7.10e-2 | 1.4× |
| | R3 | 3.84e-2 | 5.83e-2 | 5.68e-2 | 7.26e-2 | 1.02e-1 | 7.00e-2 | 1.2× |

As shown in Table S6, all 18 independently trained source models successfully support LST over all 100 target conditions, with no failed solves or non-finite predictions. For five of the six benchmarks, the source-wise mean transfer errors vary by approximately a factor of two or less. Linear ODE shows a larger relative spread, but even its least favorable source yields a mean error of only $1.09 \times 10^{-6}$; the apparent source sensitivity is therefore negligible in absolute terms.

The source-model error itself does not exhibit a simple monotonic relationship with transfer accuracy. For example, the Linear ODE source with the largest source-condition error yields one of the lowest target errors, whereas the advection source with the smallest source-condition error produces the largest mean target error. This observation complements the source-optimization analysis in Section S2.1: once a transferable response structure has formed, source-condition accuracy alone is insufficient to characterize the quality or coverage of the resulting response space.

Burgers displays the clearest source dependence in the upper tail. The median target errors remain between $9.30\times10^{-6}$ and $3.10\times10^{-5}$, demonstrating consistently accurate transfer for typical targets. In contrast, the P90 and maximum errors vary more substantially because a small number of difficult initial conditions are poorly represented by particular source response spaces. The resulting skewness also explains why the source-wise mean errors are considerably larger than their medians. Thus, typical cross-condition transfer is robust across the sampled sources, whereas coverage of rare difficult targets remains source dependent.

The trained response representation yields a lower mean target error than its matched initialization-state representation in all 18 paired comparisons. The magnitude of improvement is strongly system dependent. Burgers and Darcy exhibit gains of 30.6–82.0×and 122.1–203.0×, respectively, whereas diffusion–reaction and Navier–Stokes show smaller but consistently positive gains of 1.1–3.1×and 1.2–1.7×. The initialization-state representations nevertheless retain non-trivial recovery capacity, indicating that source-condition training strengthens an existing neural response structure rather than creating cross-condition capacity entirely from scratch. Because this comparison changes both the affine base prediction and the Jacobian-induced response modes, it measures the net effect of training on the complete response representation; the controlled ablation in Fig. 2f separately isolates these contributions for Burgers.

Taken together, these results demonstrate that cross-condition transferability is not confined to the particular initial source used in the formal ATM experiments. Independently sampled source models consistently provide usable response spaces without source-specific tuning, although the extent and uniformity of their coverage remain dependent on the governing system and source condition. This distinction is important: reproducible single-source transferability does not imply that all sources provide equivalent coverage, nor that every additional source necessarily provides complementary representation capacity. S4.2 therefore examines whether residual-guided enrichment improves coverage relative to the particular randomly sampled initial source used in each formal ATM experiment.

**S4.2 ATM performance across benchmark systems**

We next evaluate the complete ATM procedure across the six benchmark systems. Each formal experiment begins with a single randomly sampled source condition and progressively acquires new source models according to the candidate-set transfer residual. The final source library is retained when the reduction in the candidate P90 loss falls below the prescribed continuation threshold or when the maximum source budget is reached. The reductions reported in Table S7 are measured relative to the corresponding formal single-source stage and should not be interpreted as comparisons with the additional random sources examined in Section S4.1. Accordingly, this analysis evaluates the incremental benefit of residual-guided enrichment from a given initial source rather than whether the final ATM library outperforms every possible randomly selected single source.

Table S7. Final ATM performance and online memory requirements across the six benchmark systems. All losses and errors are reported at the final retained stage. Loss and error reductions are measured relative to the initial single-source stage; negative reductions indicate an increase. Memory denotes the online memory required to load the PDE-ready cache of the selected source for one target solve; caches from other retained sources need not be loaded simultaneously.

| System | Sources | Candidate P90 loss | P90 loss reduction | Mean error | Mean error reduction | Median error | P90 error | Max. error | Memory (MB) |
|---|---|---|---|---|---|---|---|---|---|
| Linear ODE | 2 | 1.24e-10 | 0.51% | 9.73e-8 | -0.51% | 7.38e-8 | 2.08e-7 | 3.40e-7 | 0.75 |
| Diffusion-reaction | 3 | 1.89e-8 | 63.12% | 8.22e-5 | 24.40% | 7.23e-5 | 1.33e-4 | 2.57e-4 | 445 |
| Burgers | 5 | 1.90e-9 | 99.86% | 1.20e-5 | 87.14% | 9.41e-6 | 2.07e-5 | 7.12e-5 | 410 |
| Advection | 3 | 6.23e-5 | 71.26% | 4.57e-3 | -0.23% | 4.24e-3 | 7.41e-3 | 1.27e-2 | 323 |
| Darcy | 3 | 5.10e-5 | 53.43% | 3.93e-4 | 39.60% | 3.74e-4 | 5.54e-4 | 6.82e-4 | 1128 |
| Navier-Stokes | 2 | 9.97e-2 | 7.15% | 4.52e-2 | -3.13% | 4.36e-2 | 5.60e-2 | 6.94e-2 | 904 |

The final source-library sizes range from two to five, showing that the residual-based stopping rule produces different enrichment depths across the benchmark systems rather than prescribing a common source budget. The clearest improvement is observed for Burgers. Successive acquisition reduces the candidate P90 loss by 99.86%, while the mean test error decreases by 87.14% relative to the formal single-source stage. The final P90 and maximum errors are $2.07\times10^{-5}$ and $7.12\times10^{-5}$, respectively, demonstrating that active enrichment substantially suppresses the difficult tail cases left by the initial response space.

Diffusion–reaction also exhibits a clear correspondence between the internal coverage metric and the external solution error. Its candidate P90 loss decreases by 63.12%, accompanied by a 24.40% reduction in mean test error and a final P90 error of $1.33\times10^{-4}$. Darcy similarly improves along its formal acquisition trajectory: the candidate P90 loss and mean test error decrease by 53.43% and 39.60%, respectively, and the final P90 error reaches $5.54\times10^{-4}$. For both systems, most of the improvement is obtained after the first actively acquired source, whereas the subsequent source produces only a marginal reduction in the candidate metric and consequently triggers stopping.

The linear ODE represents a saturation regime. Its initial response space already provides errors of order $10^{-7}$, leaving essentially no room for further improvement. The additional source changes neither the candidate loss nor the test-error distribution appreciably, and enrichment terminates after this first acquisition stage.

Advection and Navier–Stokes exhibit a common limitation of residual-guided enrichment. Although their candidate P90 losses decrease by 71.26% and 7.15%, respectively, neither system shows a corresponding reduction in mean test error. This mismatch confirms that the transfer loss is a practical residual-based coverage indicator rather than a rigorous estimator of solution error. Section S4.3 further shows that, for advection, expanding the transport-time range creates coverage gaps that ATM can successfully repair.

Overall, the results distinguish single-source transferability from active enrichability. Section S4.1 shows that reusable response spaces arise reproducibly from independently sampled source conditions, although their coverage is not uniform. ATM produces clear solution-level gains for Burgers, Darcy and diffusion–reaction, whereas the linear ODE is already saturated at the single-source stage. Advection and Navier–Stokes further show that reduced candidate residuals do not necessarily imply lower solution errors. ATM should therefore be interpreted as a residual-guided mechanism for identifying and repairing source-specific coverage gaps, rather than as a guarantee that every enrichment step improves solution accuracy or that every enriched library outperforms every randomly sampled single source.

**S4.3 Active enrichment under an expanded advection family**

The benchmark results in Section S4.2 show that active enrichment provides strongly problem-dependent benefits. For some systems, newly acquired source response spaces substantially reduce both the transfer loss and the independent solution error, whereas for others the initial source already supports broad transfer and subsequent enrichment produces little solution-level improvement. Weak enrichment does not necessarily imply that additional response spaces are intrinsically ineffective. It may instead indicate that the prescribed condition family remains largely within the effective coverage of an individual source response space, leaving only limited representational gaps for new sources to repair. Other bottlenecks, such as imperfect correspondence between the residual-based transfer loss and solution error or inaccuracies in response-coordinate evolution, may also dominate in more complex systems.

To isolate the effect of condition-domain extent, we construct a controlled extension of the advection benchmark. The advection equation is particularly suitable for this analysis because the induced solution family can be expanded through a single physically interpretable transport-scale parameter while retaining the governing equation, stochastic coefficient structure, initial and boundary conditions, network architecture and LST configuration. Specifically, the transport coefficient is defined as $a(x;c,g)=ca_0(x;g)=c[g(x)-\min_{\xi\in[0,1]} g(\xi)+1]$, where $g\sim\mathcal{GP}(0,K), c\sim\mathcal{U}(0.1,1)$. The original advection benchmark corresponds to $c=1$. Because $a_0(x;g)$ is strictly positive, the multiplicative factor preserves the direction of transport while varying its global time scale. Representative solutions obtained from the same Gaussian-process realization are shown in Fig. S7. Decreasing c slows characteristic propagation, thereby expanding the solution family across distinct transport time scales.

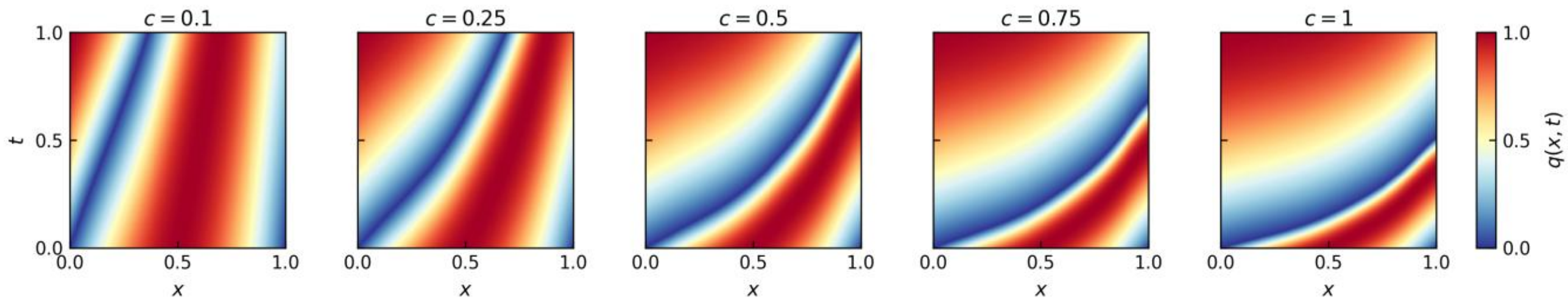

Fig. S7. Advection solutions at different global transport scales.

The initial source is the source model used in the original formal advection experiment and is retained at $c=1$ without retraining. The candidate and test sets each contain 100 conditions, with the Gaussian-process realizations and scale factors sampled independently between the two sets. All remaining settings are identical to those of the original advection benchmark. No source- or target-specific tuning is performed.

ATM progressively reduces the candidate upper-tail loss over the expanded condition family (Table S8). The first acquired source lies in the low-transport-scale region at $c=0.255$ and produces the largest single-stage reduction, whereas subsequent acquisitions refine coverage across low, intermediate and near-original transport scales.

Table S8. Stagewise ATM construction over the expanded advection condition family. The newly acquired source scale denotes the global coefficient multiplier added at each stage. The marginal reduction is the relative decrease in candidate P90 loss from the preceding stage. Test errors are relative $L_2$ errors evaluated on the same 100 independent

test conditions.

| Stage | Newly acquired $c$ | Candidate P90 loss | Marginal reduction | Mean test error | P90 test error |
|---|---|---|---|---|---|
| 1 | 1.000 | 4.27e-4 | — | 9.06e-3 | 1.50e-2 |
| 2 | 0.255 | 1.35e-4 | 0.68 | 4.72e-3 | 7.33e-3 |
| 3 | 0.363 | 1.17e-4 | 0.14 | 4.54e-3 | 7.32e-3 |
| 4 | 0.628 | 6.81e-5 | 0.42 | 4.29e-3 | 6.92e-3 |
| 5 | 0.626 | 5.12e-5 | 0.25 | 4.33e-3 | 6.88e-3 |
| 6 | 0.155 | 3.49e-5 | 0.32 | 4.34e-3 | 7.00e-3 |
| 7 | 0.822 | 2.80e-5 | 0.20 | 3.78e-3 | 5.70e-3 |
| 8 | 0.958 | 2.01e-5 | 0.28 | 3.77e-3 | 5.70e-3 |
| 9 | 0.360 | 1.92e-5 | 0.04 | 3.63e-3 | 5.69e-3 |

Under the fixed stopping threshold used throughout this study, ATM terminates at stage 9 because the final marginal reduction in candidate P90 loss falls below 0.10. Relative to the initial single-source stage, the final candidate P90 loss decreases by 0.96, while the mean and P90 test errors decrease by 0.60 and 0.62, respectively. The small non-monotonic variations in test error at intermediate stages are expected because acquisition is driven exclusively by the candidate residual rather than by reference-solution error. The repeated selection of sources with similar values of $c$, such as those near 0.63 and 0.36, further indicates that source complementarity depends on the spatial realization of the coefficient field as well as its global scale.

As in the Burgers experiment, most solution-level improvement occurs before the final enrichment stages. The first acquired source accounts for approximately 80% of the total reduction in both mean and P90 test errors, and by stage 7 the P90 error is already nearly identical to its final stage-9 value. The nine-source library therefore reflects the conservative fixed stopping threshold used here rather than a minimum source requirement for achieving most of the accuracy gain. In applications with higher source-construction costs, the threshold or source budget may be adjusted to obtain an earlier operating point on this cost–accuracy trade-off. Because the marginal candidate-loss reductions are not monotonic, such adjustment should be interpreted as an application-dependent trade-off rather than a universally preferable stopping rule.

To determine whether these improvements reflect genuinely expanded representation capacity, we evaluate the original $c=1$ source response space over five fixed transport scales. Table S9 compares its oracle affine projection error with the stage-1 LST error and the final nine-source ATM error, using the same 100 Gaussian-process realizations at each scale.

Table S9. Fixed-scale diagnosis of response-space coverage. Projection denotes the oracle affine projection onto the original $c$ = 1 source response space. Stage 1 uses only this source, whereas final ATM uses the retained nine-source library.

| $c$ | Projection mean | Projection P90 | Stage-1 mean | Stage-1 P90 | Final ATM mean |
|---|---|---|---|---|---|
| 0.10 | 9.46e-3 | 1.13e-2 | 1.44e-2 | 1.76e-2 | 2.09e-3 |
| 0.25 | 7.37e-3 | 1.11e-2 | 1.20e-2 | 1.78e-2 | 2.70e-3 |
| 0.50 | 4.57e-3 | 6.40e-3 | 9.43e-3 | 1.55e-2 | 4.80e-3 |
| 0.75 | 3.10e-3 | 4.46e-3 | 6.28e-3 | 9.97e-3 | 3.47e-3 |
| 1.00 | 2.42e-3 | 3.90e-3 | 4.56e-3 | 7.20e-3 | 3.48e-3 |

The oracle projection analysis shows that the original source space loses coverage as the transport scale departs from its training value. ATM produces its largest gains in this weak-coverage regime, with the final mean errors falling below the oracle projection floor of the original affine space in the lowest-scale, weak-coverage regime. This demonstrates that the acquired sources provide genuinely new affine representation capacity rather than merely improving coordinate optimization within the initial source space. These results clarify the distinction between transferability and enrichability. A single source may support broad transfer over a comparatively compact condition family, leaving little opportunity for enrichment. Once the condition domain induces solution variations beyond that coverage, however, residual-guided acquisition can identify complementary response spaces and substantially improve both typical and upper-tail accuracy. This controlled experiment establishes condition-domain extent as one mechanism governing the problem-dependent enrichment behavior observed across the benchmark suite, although other systems may additionally be limited by residual–error mismatch or response-coordinate evolution.